\documentclass[journal]{IEEEtran}

\usepackage{booktabs}
\usepackage{graphicx}
\usepackage{amsmath}
\usepackage{array}
\usepackage{arydshln}
\usepackage[utf8]{inputenc}
\usepackage[french]{nomencl}
\usepackage{subfigure}
\usepackage{amssymb}
\usepackage{xcolor}
\usepackage{nomencl}
\usepackage{cite}
\usepackage{mathrsfs}
\usepackage{cases}
\usepackage{float}
\usepackage[ruled,vlined]{algorithm2e}
\usepackage[capitalise]{cleveref}
\usepackage{booktabs}
\usepackage{siunitx}
\makenomenclature

\newcommand{\xun}[1]{{\color{black}#1}}
\newcommand{\xunn}[1]{{\color{black}#1}}

\ifCLASSINFOpdf

\else

\fi

\begin{document}

%
\title{CT-Enabled Patient-Specific Simulation and Contact-Aware Robotic Planning for Cochlear Implantation}
%
%
%

\author{Lingxiao Xun, Gang Zheng, Alexandre Kruszewski, Renato Torres

\thanks{ }
}
%
%

\markboth{}%
{Shell \MakeLowercase{\textit{et al.}}: Bare Demo of IEEEtran.cls for IEEE Journals}
%



\maketitle

\begin{abstract}
Robotic cochlear-implant (CI) insertion requires precise prediction and regulation of contact forces to minimize intracochlear trauma and prevent failure modes such as locking and buckling. Aligned with the integration of advanced medical imaging and robotics for autonomous, precision interventions, this paper presents a unified CT-to-simulation pipeline for contact-aware insertion planning and validation. We develop a low-dimensional, differentiable Cosserat-rod model of the electrode array coupled with frictional contact and pseudo-dynamics regularization to ensure continuous stick–slip transitions. Patient-specific cochlear anatomy is reconstructed from CT imaging and encoded via an analytic parametrization of the scala-tympani lumen, enabling efficient and differentiable contact queries through closest-point projection. Based on a differentiated equilibrium-constraint formulation, we derive an online direction-update law under an RCM-like constraint that suppresses lateral insertion forces while maintaining axial advancement. Simulations and benchtop experiments validate deformation and force trends, demonstrating reduced locking/buckling risk and improved insertion depth. The study highlights how CT-based imaging enhances modeling, planning, and safety capabilities in robot-assisted inner-ear procedures.
\end{abstract}


%
\IEEEpeerreviewmaketitle

\section{Introduction}

Robot-assisted cochlear-implant (CI) insertion is an image-guided intervention in which a compliant electrode array
(EA) must be advanced through a highly confined lumen while regulating contact forces to minimize intracochlear trauma
and to prevent failure modes such as locking and buckling
\cite{jwair2021scalar,dhanasingh2019review}. A key difficulty is that
electrode--lumen interaction is strongly shaped by patient anatomy: small variations in lumen curvature and
cross-sectional morphology can shift contact locations and substantially change the resulting force profile and
failure onset. Consequently, patient-specific anatomy derived from preoperative imaging is not merely descriptive but
directly impacts the fidelity of simulation and the reliability of planning.

Computed tomography (CT) is routinely available in CI workflows and provides high-resolution information of the bony
labyrinth, making it a practical source for reconstructing the scala-tympani lumen geometry. However, directly using
CT-derived surface meshes for contact-rich simulation and planning can be computationally expensive and may hinder
sensitivity-based updates due to non-smooth geometric queries. This motivates anatomy representations that remain
patient-specific while supporting efficient contact computation, including closest-point queries, surface normals,
and penetration measures, and, when needed, differentiability for optimization and control.

The following subsections review prior work on mechanical and contact modeling of CI insertion, CT-based cochlear
anatomy reconstruction and compact lumen representations, and image-guided planning and robotic insertion strategies.

\subsection{Related work on modeling: electrode mechanics, CT-derived anatomy, and contact}

\subsubsection{Electrode-array mechanics models}
Mechanical modeling of cochlear implants, and in particular the EA, lies at the intersection of continuum mechanics
and medical-device design. Owing to its slender and highly compliant geometry, the EA is naturally modeled as a
beam/rod-like continuum structure. General-purpose finite element methods (FEM) can capture complex geometries and
material behaviors \cite{elsayed2014finite}, yet full 3D FEM discretizations are often unnecessarily expensive for
long, slender devices whose dominant deformation modes can be represented with far fewer degrees of freedom. Beam and
rod models address this mismatch by exploiting the slender-body assumption, offering an efficient description for CI
insertion. Classical Euler--Bernoulli beam theory provides a baseline for quasi-static bending; Olson et al. adopted a
quasi-static bending model to study EA mechanics and derive design-relevant insights \cite{olson2020euler}.

When large deformation, torsion, and shear become important, constant-curvature assumptions and Cosserat-rod models
are commonly used. Cosserat rod theory, as a geometrically exact generalization of Timoshenko--Reissner beams,
captures bending, torsion, shear, and extension in a unified framework
\cite{boyer2017poincare,cao2008nonlinear}. Till et al. reviewed Cosserat-based models for slender
elastic rods, including numerical strategies relevant to CI modeling \cite{till2019statics}. Continuous-space solvers
based on Newton--Euler dynamics have also been developed to solve Cosserat PDEs efficiently \cite{till2019real}. To
reduce computational burden further, \cite{8500341} introduced a piecewise-constant strain representation and derived
an approximate weak form that transforms the governing PDEs into a low-dimensional ODE system. Related discrete
Cosserat formulations based on strain parameterization and Lagrangian dynamics have been proposed in \cite{boyer2020dynamics}.
Despite these advances, accurately capturing implantation behaviors in the presence of contact, including local strain
variations and buckling phenomena, remains challenging, as contact introduces strong nonlinearities and rapidly
varying loads; this motivates low-dimensional formulations that remain robust in contact-rich regimes and can support
planning-oriented sensitivity analysis.

\subsubsection{CT-derived cochlear anatomy models for simulation and planning}
In parallel to implant mechanics, cochlear modeling has been approached at different levels of geometric fidelity,
ranging from high-resolution patient-specific surface/volume reconstructions to compact analytic descriptions.
CT- or $\mu$CT-based segmentation followed by mesh/FEM model construction is widely used when anatomical accuracy is
critical, for instance in patient-specific in-silico assessment of implant placement and stimulation
\cite{ceresa2014patient,mangado2018towards}. For robotics and optimization, however, mesh-based representations can be
costly for repeated geometric queries and may complicate differentiation, since operations such as closest-point
queries and normal evaluation are not always smooth on discrete surfaces. Several works therefore exploit the tubular
nature of the scala tympani by representing the lumen through a centerline coupled with a parametric cross-section,
yielding swept-tube or low-order models suitable for simulation, planning, and experimental phantoms
\cite{salkim2022insertion,schurzig2021uncoiling,thiselton2024parameterisation}. These representations highlight an
important design space: anatomy models should be faithful to CT-derived patient variability while remaining compact
enough to support fast, repeated geometric queries and, when needed, differentiable contact computations for
optimization and control.

\subsubsection{Contact modeling for cochlear implantation}
The interaction between the EA and the cochlear lumen is typically modeled using contact mechanics. Contact modeling
for cochlear implantation is particularly demanding due to the combination of large deformation, geometric complexity,
and frictional effects \cite{schegg2022review}. Existing simulators have explored contact-aware cochlear implantation
in different contexts. Goury et al. developed a patient-specific numerical simulation tool within the SOFA framework
to model 3D insertion and to support virtual planning and robot-assisted procedures \cite{goury2016numerical}. Jones
et al. presented a virtual-reality simulator that reproduces electrode behavior during insertion for surgical training
\cite{jones2019virtual}.

From a numerical standpoint, two families of contact formulations are commonly used. Penalty methods model contact
forces through compliant regularization and are simple to implement, but their accuracy depends on stiffness tuning
and may allow non-physical penetration \cite{wriggers2006computational}. Complementarity-based formulations enforce
non-penetration and frictional constraints more precisely, for example via linear or nonlinear complementarity
problems, but they introduce inequality constraints and non-smoothness that complicate controller design and
differentiation \cite{anitescu1997formulating}. This motivates contact models that remain physically meaningful while
being numerically robust and compatible with sensitivity-based planning and online updates.

\subsection{Related work on planning and control: image-guided trajectories and interaction-aware insertion}
Path planning for CI insertion aims to identify an approach direction and an insertion motion that (1) enables reliable
access to the scala tympani and (2) reduces intracochlear contact forces to mitigate trauma and buckling risk. Early
work on robot-assisted insertion highlighted that adapting the approach direction and steering the electrode can
markedly affect insertion forces, motivating planning and control beyond constant-path insertion. Zhang et al.
formulated optimal path-planning problems for robotic insertion of steerable electrode arrays and showed that
increasing insertion degrees of freedom and adapting the approach direction can reduce insertion forces in simulation
and experiments \cite{zhang2009optimal,zhang2008path}.

A complementary line of research focuses on image-guided trajectory design based on patient anatomy. CT-derived models
have been used to determine subject-specific access trajectories that satisfy anatomical constraints and account for
drill/tool positioning errors, thereby providing a feasible surgical corridor \cite{noble2007determination,noble2010automatic}.
Such image-based planning is essential for defining a safe initial approach, yet it typically does not model the
contact-rich mechanics during intracochlear insertion, where frictional interaction and lumen curvature dominate the
force profile and the resulting EA path. This gap suggests that image-guided planning benefits from being coupled with
mechanics-based interaction prediction inside the cochlea.

To incorporate interaction, several approaches rely on steering or actuation mechanisms to modulate insertion
direction and reduce contact forces. Magnetically steered robotic insertion, for instance, demonstrated the benefit of
direction adaptation with system-level integration and experimental validation \cite{bruns2020magnetically}. These
results suggest that effective planning should be coupled with an interaction model capable of predicting and
regulating contact forces in patient-specific anatomy, and that such a model should be efficient enough to support
iterative updates.

From a surgical-robotics perspective, insertion is performed through an anatomical entry point, which motivates a
remote-center-of-motion (RCM) constraint or RCM-like constraint to keep the tool shaft pivoting about the entry
while allowing controlled axial advancement. RCM constraints and, more broadly, constraint-based surgical motion
planning have been widely used to enforce safe entry-point kinematics and to combine task execution with virtual
safety constraints \cite{kuntz2017motion,marinho2019dynamic}. In the context of CI insertion, this indicates that
planning should not be treated purely as an offline geometric problem, but rather as an online direction-update and
control problem under an RCM-like constraint, informed by contact interaction inside the cochlea.

Overall, existing studies motivate patient-specific, image-derived models that couple cochlear anatomy and electrode
mechanics in a contact-aware manner, while remaining compatible with sensitivity analysis for online direction
adaptation under clinically motivated constraints.

\subsection{Contributions and outline}
This paper contributes a unified CT-to-simulation pipeline for contact-aware robotic insertion planning and
validation, by coupling a low-dimensional differentiable simulator with an RCM-like insertion constraint to enable
online direction adaptation. Specifically, we:
\begin{enumerate}
\item develop a low-dimensional contact-mechanics model for EA insertion based on a Cosserat-rod formulation, together
with a patient-specific, analytic, and differentiable parametrization of the scala-tympani lumen for efficient
closest-point queries (and associated normals/penetration measures);
\item derive a continuous-time, contact-aware direction-update law under an RCM-like constraint by differentiating the
resulting equilibrium--constraint system, and validate the model and the resulting planning strategy in both
simulation and experiments.
\end{enumerate}

The remainder of this paper is organized as follows. Section~\ref{section:prob} presents the problem statement.
Section~\ref{section:modeling} details the Cosserat-based EA model and the contact-mechanics formulation.
Section~\ref{section:geometric} introduces the analytic cochlear lumen parametrization. The insertion dynamics and the
proposed contact-aware direction update law are detailed in Sec.~\ref{section:dynamics} and Sec.~\ref{section:control}.
Section~\ref{section:experiment} and Sec.~\ref{section:experiment2} report simulation and experimental validations.
Section~\ref{section:conclusion} concludes the paper and outlines future directions.

\section{Problem statement}\label{section:prob}
\begin{figure}[t]
	\centering
	\includegraphics[width=0.48\textwidth]{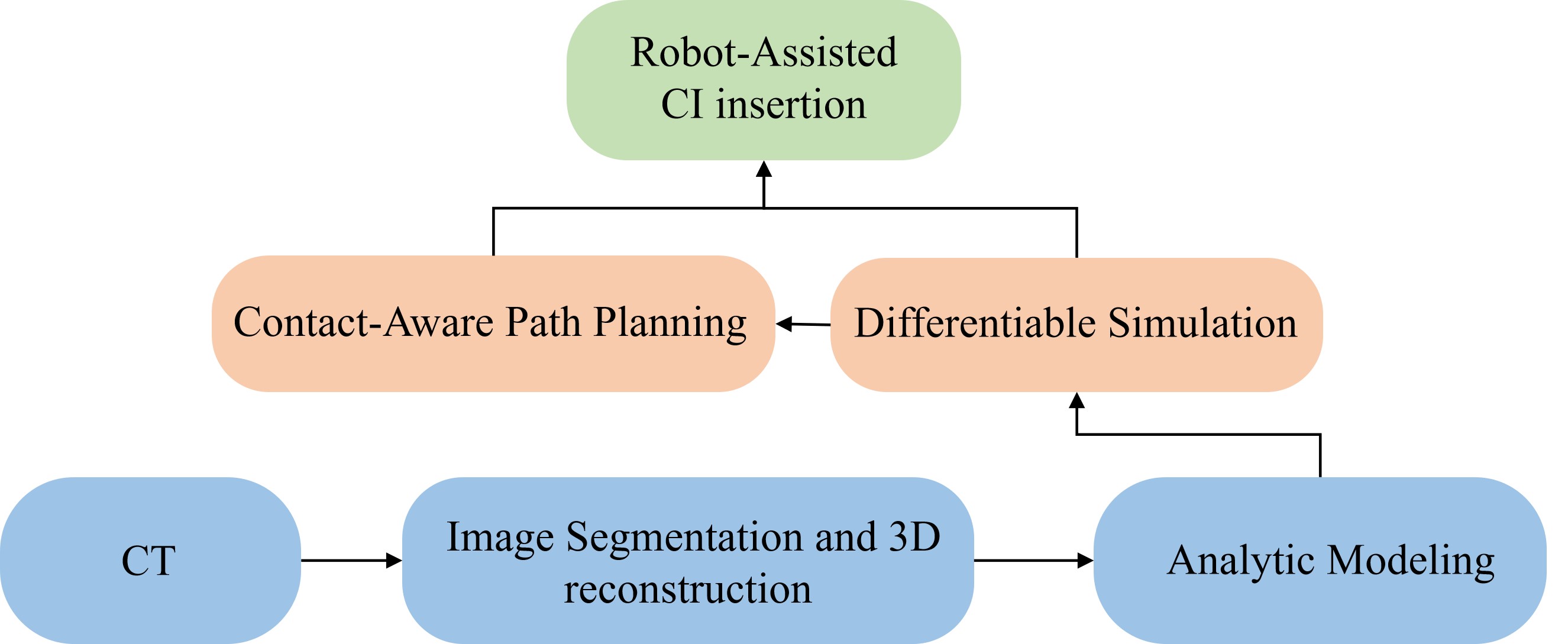}
	\caption{CT-to-simulation pipeline for contact-aware insertion planning.}
	\label{fig:pipline}
\end{figure}

Advanced imaging techniques such as CT, together with accurate segmentation and 3D reconstruction, provide patient-specific anatomical information that is critical for robot-assisted cochlear implantation (CI). Fig.~\ref{fig:pipline} overviews the CT-to-simulation framework proposed in this paper: the imaging-derived anatomy is converted into an analytic cochlear model, coupled with differentiable simulations of electrode-array mechanics, and integrated with contact-aware path planning for predictive, CT-enabled insertion planning and execution. This pipeline improves insertion precision while monitoring interaction forces and reducing the risk of trauma.

More broadly, the proposed CT-to-simulation pipeline illustrates how medical imaging can directly inform modeling, simulation, and control, thereby supporting autonomous and safety-aware CI insertion strategies. Since CT imaging and segmentation are mature and widely available in clinical practice, it is timely to demonstrate how these tools can be leveraged for CI modeling, simulation, and control, which constitute the main focus of the remainder of this paper. Achieving this goal requires repeatedly evaluating a physics-based model under varying base orientations and performing reliable gradient-based updates during optimization, which leads to three coupled technical challenges.

First, the electrode array (EA) undergoes large deformation and interacts with the cochlear wall through distributed
contact and friction. A mechanically consistent EA model is required to capture its deformation during insertion and
to expose the base configuration as the control input for planning.

Second, the cochlea is patient-specific and must be reconstructed from CT/$\mu$CT imaging. For simulation and
planning, the cochlear lumen model should satisfy three key requirements: (i) it must be patient-specific to capture
inter-subject variability in lumen morphology; (ii) it should admit a compact parameterization so that geometric
contact queries (closest-point search, normal evaluation, penetration measures) can be performed rapidly within
repeated simulation loops; and (iii) it should provide stable local geometric derivatives with respect to spatial
coordinates and model parameters, as required by sensitivity analysis and gradient-based updates.

Third, contact mechanics is intrinsically nonsmooth, yet path optimization and online direction adaptation require a
differentiable (or sensitivity-enabled) contact formulation to compute consistent gradients of insertion forces with
respect to the control variables. Without such a formulation, iterative updates become unreliable or prohibitively
slow.

Motivated by these challenges, this paper addresses the following questions:
\begin{enumerate}
	\item How can we build an efficient mechanical model of the EA suitable for insertion simulation and for defining
	base-level control variables?
	\item Based on CT/$\mu$CT imaging, how can we construct a patient-specific cochlear lumen model that supports fast
	contact computation and provides geometric derivatives for subsequent sensitivity analysis?
	\item How can we formulate a differentiable (sensitivity-enabled) contact mechanics model and leverage it to solve a
	path-optimization problem under clinically meaningful kinematic constraints?
\end{enumerate}

We begin by introducing the EA model to clarify the mechanical variables and the required contact quantities. We then
present the CT-derived cochlear geometry model, followed by the differentiable contact formulation and the resulting
simulation-guided path-optimization method and its validation.

\section{Modeling of Implant}\label{section:modeling}
Cochlear implantation is a contact-rich insertion process in which a slender and highly compliant electrode array (EA) interacts with the patient-specific cochlear lumen. For simulation-guided path planning in robot-assisted procedures, the mechanics model must capture the EA's large deformations and provide a consistent way to compute contact forces and their sensitivities with respect to the base configuration, which corresponds to robot-controllable insertion parameters. To ensure patient specificity while keeping repeated simulations tractable for gradient-based optimization or online adaptation, we later introduce in Sec.~\ref{section:geometric} an analytic cochlear lumen geometry derived from CT imaging, enabling efficient and differentiable geometric queries for contact handling.

Beam models are well suited for slender structures, and the Cosserat-rod formulation provides a geometrically exact description of large rotations and strains. In this section, we present the Cosserat-rod model used for the EA and formulate the contact mechanics that determine the external loads in the governing equations, establishing a CT-informed and contact-aware foundation for safety-conscious insertion planning and robotic control.

\begin{figure}[t]
	\centering
	\includegraphics[width=0.45\textwidth]{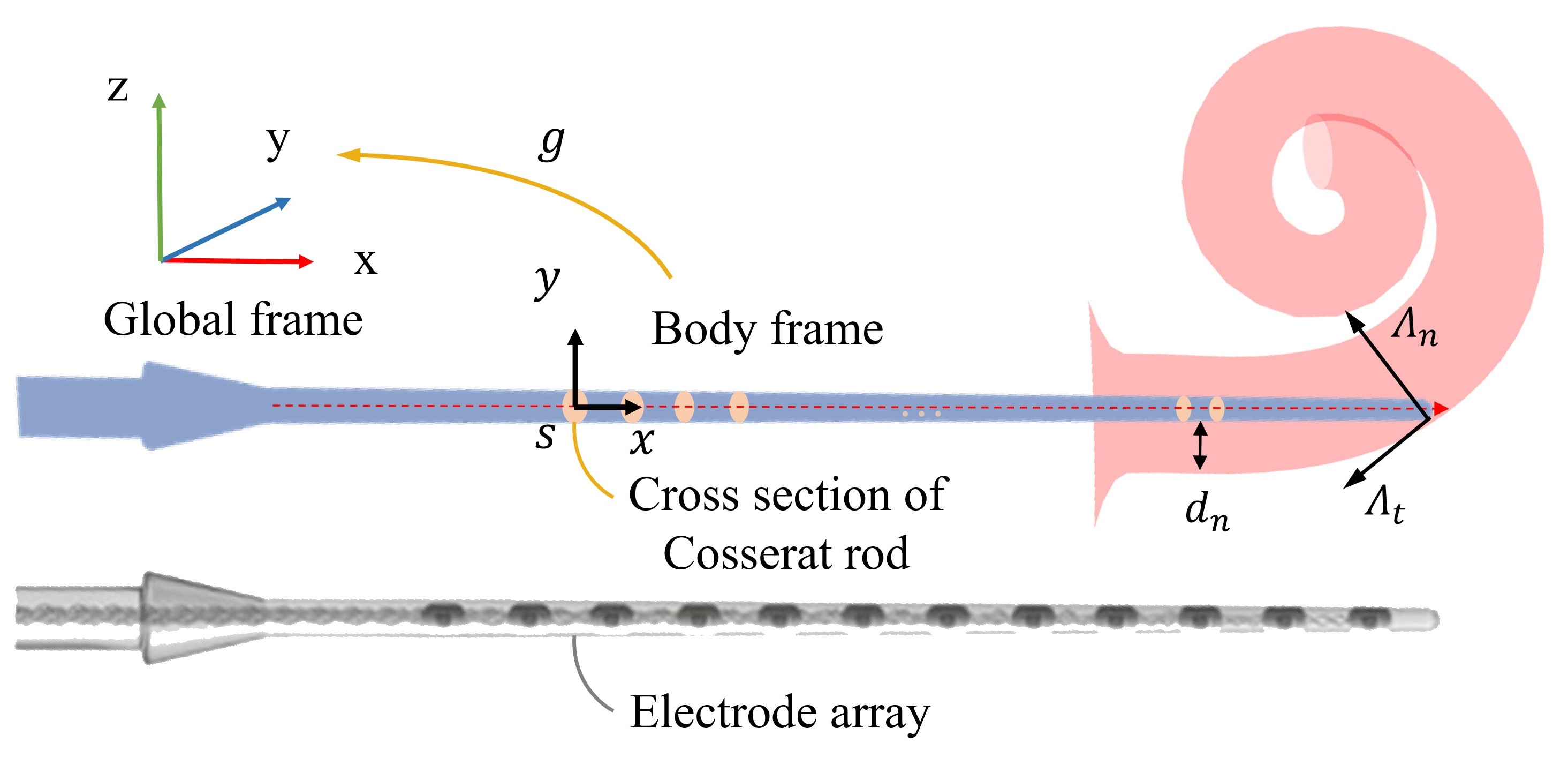}
	\caption{Cosserat-rod model of the implant electrode array (EA).}
	\label{fig:implant}
\end{figure}
\subsection{Cosserat modeling}\label{sec:1}
As shown in Fig.~\ref{fig:implant}, the centerline of the implant (EA) can be parameterized by the arc-length coordinate $s$, with position $\boldsymbol{p}(s)\in\mathbb{R}^3$ and cross-sectional orientation $\boldsymbol{R}(s)\in SO(3)$. The configuration of each cross-section is represented by the homogeneous transformation
\begin{equation}\label{gmatrix}
	\boldsymbol{g}(s)=\begin{bmatrix}
	\boldsymbol{R}(s) & \boldsymbol{p}(s) \\
	\boldsymbol{0}^{\top} & 1
\end{bmatrix}\in SE(3).
\end{equation}
For simplicity, we use $(\cdot)^{\prime}$ to denote $\partial(\cdot)/\partial s$ and $\dot{(\cdot)}$ to denote $\partial(\cdot)/\partial t$.
The body strain and body velocity of the cross-section are defined, respectively, as
\[
\boldsymbol{\xi}=(\boldsymbol{g}^{-1}\boldsymbol{g}^{\prime})^{\vee}\in \mathbb{R}^{6},\qquad
\boldsymbol{\eta}=(\boldsymbol{g}^{-1}\dot{\boldsymbol{g}})^{\vee}\in \mathbb{R}^{6}.
\]
where $(\cdot)^\vee$ denotes the standard mapping from $\mathfrak{se}(3)$ to $\mathbb{R}^6$. 
The corresponding kinematic equations can be written as \cite{8500341}:
\begin{equation}\label{kincos}
	\frac{\partial}{\partial s}
	\begin{pmatrix}
		\boldsymbol{g}\\ \boldsymbol{\eta}\\ \dot{\boldsymbol{\eta}}
	\end{pmatrix}
	=
	\begin{pmatrix}
		\boldsymbol{g} \widehat{\boldsymbol{\xi}}\\
        \dot{\boldsymbol{\xi}}-\operatorname{ad}_{\boldsymbol{\xi}} \boldsymbol{\eta}\\
		\ddot{\boldsymbol{\xi}}-\operatorname{ad}_{\dot{\boldsymbol{\xi}}} \boldsymbol{\eta}-\operatorname{ad}_{\boldsymbol{\xi}} \dot{\boldsymbol{\eta}}
	\end{pmatrix},
\end{equation}
together with the dynamic balance \cite{boyer2020dynamics}:
\begin{equation}\label{contactpde0}
	\boldsymbol{\mathcal{M}}\dot{\boldsymbol{\eta}}
    -\operatorname{ad}_{\boldsymbol{\eta}}^{\top}\boldsymbol{\mathcal{M}}\boldsymbol{\eta}
    =
    \boldsymbol{\Lambda}_{i}^{\prime}
    -\operatorname{ad}_{\boldsymbol{\xi}}^{\top} \boldsymbol{\Lambda}_{i}
    +{\boldsymbol{\Lambda}}_{e},
\end{equation}
where $\boldsymbol{\mathcal{M}}\in \mathbb{R}^{6\times6}$ is the generalized mass density, $\boldsymbol{\Lambda}_{i}\in \mathbb{R}^{6}$ is the internal wrench, and ${\boldsymbol{\Lambda}}_{e}\in \mathbb{R}^{6}$ denotes the external wrench (e.g., contact loads). The operator $\operatorname{ad}_{(\cdot)}$ is defined in Appendix~\ref{notations}.
Equation~\eqref{contactpde0} governs the EA dynamics during insertion; the remaining challenge is to determine the contact contribution ${\boldsymbol{\Lambda}}_{e}$ in a way that is efficient and compatible with simulation-guided optimization.

Beyond the intrinsic deformation of the EA, its interaction with the cochlear wall must be modeled through contact forces and constraints.
Following \cite{10494907}, we impose a unilateral normal constraint (Signorini condition) together with Coulomb friction.
As shown in Fig.~\ref{fig:implant}, let $d_n$ denote the signed normal gap between a potential contact pair (positive in separation) and let $\Lambda_n$ denote the corresponding normal contact force.
The normal contact condition reads
\begin{equation}\label{nncp111}
	0\leq d_n\ \bot\ \Lambda_{n}\geq 0.
\end{equation}
Let $\boldsymbol{v}_{t}$ be the relative tangential slip velocity and $\boldsymbol{\Lambda}_{t}$ the tangential friction force. Coulomb friction is enforced by
\begin{equation}\label{coulomb_cone}
	\|\boldsymbol{\Lambda}_{t}\|\leq \mu \Lambda_{n},
\end{equation}
together with the maximum-dissipation/sliding relation when $\boldsymbol{v}_{t}\neq \boldsymbol{0}$,
\begin{equation}\label{coulomb_slide}
	\boldsymbol{\Lambda}_{t} = -\mu \Lambda_{n}\frac{\boldsymbol{v}_{t}}{\|\boldsymbol{v}_{t}\|}.
\end{equation}
Here, $\mu$ is the friction coefficient and the symbol ``$\bot$'' indicates complementarity (i.e., $\delta_n\,\Lambda_n=0$).
These contact constraints, together with the rod dynamics \eqref{contactpde0}, define the insertion mechanics.

Enforcing \eqref{nncp111}--\eqref{coulomb_slide} requires identifying potential contact pairs between the EA and the
cochlear lumen and evaluating local geometric quantities (closest points, normals, and gap functions) in a
consistent manner. Due to the highly curved and patient-specific cochlear anatomy, repeated closest-point queries on
CT-derived surface meshes may become a computational bottleneck in simulation and path-optimization loops. This
motivates the analytic cochlear lumen geometry model introduced in the next section, which enables fast and
differentiable contact queries to support sensitivity-enabled simulation for path optimization.

\section{Image-Derived Geometric Modeling of the Cochlear Lumen}\label{section:geometric}

For contact-rich CI insertion simulation and sensitivity-based path optimization, the cochlear lumen model must enable
repeated evaluation of local geometric quantities required by contact, including closest points, surface normals, and
gap functions, and provide stable local derivatives when gradients with respect to the insertion configuration are
needed. While patient-specific cochlear anatomy can be reconstructed from CT/$\mu$CT imaging, the resulting
high-resolution surface meshes may become computationally costly in iterative simulation and optimization loops, and
mesh-based closest-point and normal evaluations can be numerically non-smooth around discretization artifacts. These
considerations motivate a compact analytic geometry model that remains patient-specific while supporting fast and
differentiable contact queries.

In this work, the cochlea is modeled as a rigid anatomical cavity, i.e., we neglect any deformation of the bony
labyrinth during implantation. Geometrically, the scala-tympani lumen is represented as a swept tube generated by
sweeping a varying cross-section along a three-dimensional centerline with an associated local moving
frame. Accordingly, the analytical construction is organized into three steps:
\begin{enumerate}
  \item extract and parameterize the cochlear centerline,
  \item define a local moving frame along the centerline,
  \item specify the cross-sectional profile in each local frame.
\end{enumerate}
The details are presented below.

\subsection{Overview of the cochlear parametrization}\label{sec:param_overview}

Patient-specific scala tympani (ST) geometries are typically obtained from CT-based segmentation and represented as
high-resolution triangular meshes. While accurate, such meshes make contact-mechanics simulation of cochlear-implant
(CI) insertion expensive, as contact detection requires repeated closest-point queries and contact-pair generation on
dense, non-smooth discretized surfaces.

To reduce this cost, we adopt a low-dimensional, analytic, and differentiable parametrization of the ST lumen. As shown in Fig.~\ref{fig:cochleamodel5},
we represent the ST as a tubular surface driven by a Cosserat-rod pose curve on $SE(3)$,
\begin{equation}
\boldsymbol{g}_c(s)=
\begin{bmatrix}
\boldsymbol{R}_c(s) & \boldsymbol{r}(s)\\
0 & 1
\end{bmatrix}\in SE(3),\qquad s\in[0,L],
\end{equation}
where $\boldsymbol{r}(s)\in\mathbb{R}^3$ is the centerline and $\boldsymbol{R}_c(s)\in SO(3)$ is an attached orthonormal frame. The longitudinal
parameter $s$ is chosen as the centerline arc length, and a circumferential parameter $\beta\in[0,2\pi)$ specifies
locations on the cross-section around the local tangential axis. More generally, the cross-sectional contour is
described by a parametric function $\boldsymbol{f}(s,\beta)$; its specific form will be introduced in the following sections.
Together, $\boldsymbol{g}_c(s)$ and $\boldsymbol{f}(s,\beta)$ define a smooth two-parameter surface and thus an analytic ST geometry.

This representation (i) enables faster contact computations on a continuous surface (or with controlled sampling),
(ii) provides a compact interface consistent with rod-based CI models, and (iii) facilitates rapid morphometric
adjustments when CT data are noisy or unavailable, thereby accelerating the overall simulation workflow.

\subsection{From CT volumes to a 3D surface mesh}\label{sec:ct_to_mesh}

\begin{figure}[t]
	\centering
	\includegraphics[width=0.49\textwidth]{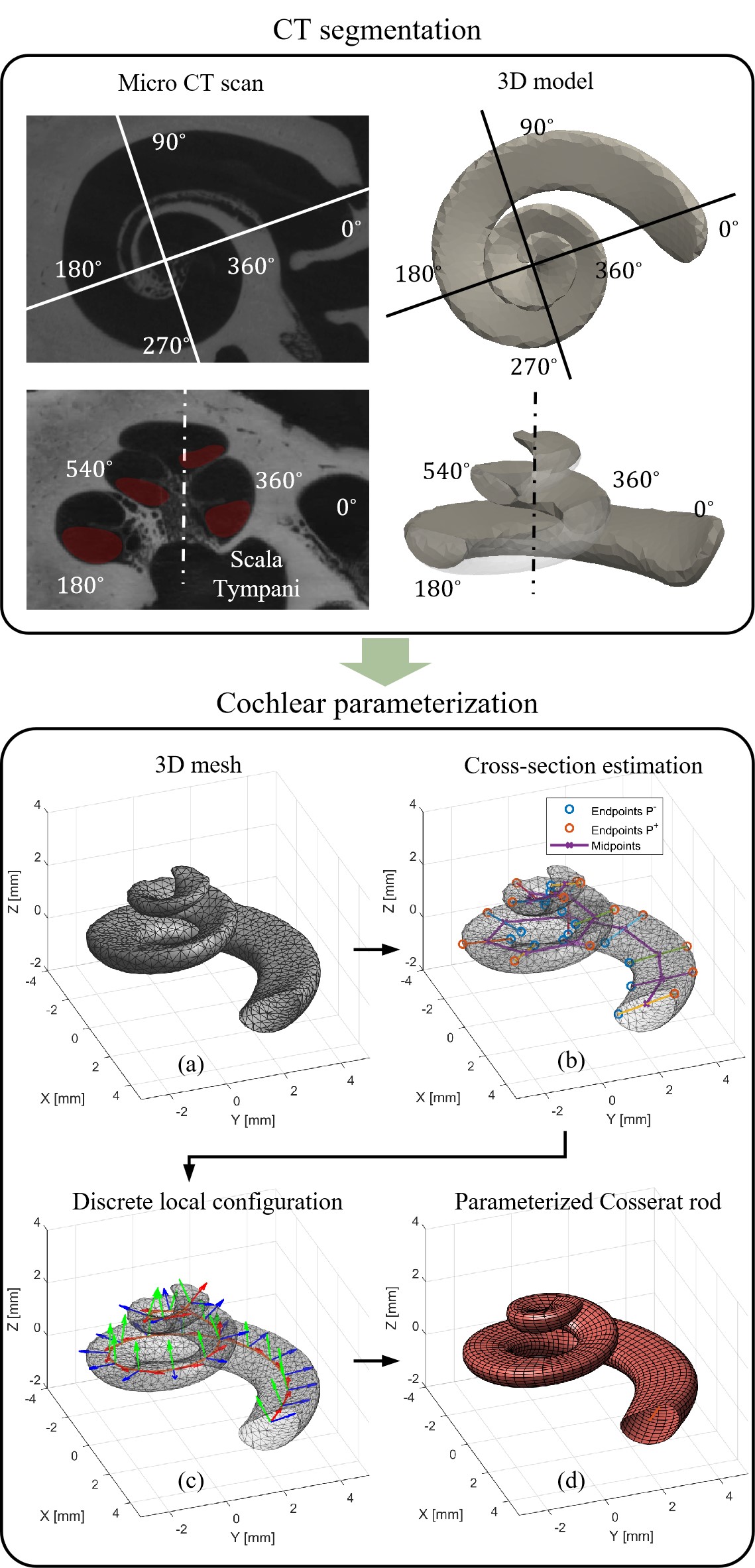}
    \caption{Patient-specific cochlear lumen modeling pipeline: CT segmentation $\rightarrow$ 3D surface mesh $\rightarrow$ cross-section estimation and discrete local frames $\rightarrow$ piecewise-constant-strain Cosserat-rod parameterization.}
	\label{fig:scanmodel}
\end{figure}

We start from clinical CT scans provided as DICOM volumes. The image data are first handled using the
Insight Segmentation and Registration Toolkit (ITK)~\cite{mccormick2014itk} to reliably import the volumetric data and perform basic
pre-processing (e.g., region-of-interest extraction and format conversion) when needed. The resulting volume is then
used to obtain a patient-specific representation of the scala tympani (ST) lumen via semi-automatic segmentation, as illustrated in Fig.~\ref{fig:scanmodel}.

The segmented lumen is converted into a triangular surface mesh $\mathcal{M}$ (e.g., exported as an STL surface),
which serves as the geometric boundary of the ST for the subsequent steps. Throughout the pipeline, we preserve
physical units (millimeters) and enforce a consistent coordinate system between the exported mesh and all derived
geometric entities.

Because CT resolution and segmentation artifacts may locally affect the reconstructed surface, the following sections
avoid relying on fine mesh details and instead extract robust low-dimensional descriptors (centerline and
cross-sectional parameters) through smoothing and statistical estimation.

\subsection{Centerline extraction and automated cross-section estimation}\label{sec:centerline_and_sections}

Given the reconstructed scala-tympani lumen surface mesh $\mathcal{M}$, we extract a compact set of geometric
descriptors consisting of (i) a centerline curve and (ii) cross-sectional parameters evaluated at a set of
arc-length locations. Importantly, the sampling along the centerline is not restricted to uniform spacing: the set
$\{s_i\}_{i=0}^{N}$ can be chosen uniformly or refined in regions of interest (e.g., high curvature or rapid
cross-sectional variation), enabling an efficient trade-off between geometric fidelity and computational cost.

\subsubsection{Centerline extraction and arc-length parametrization.}
A centerline is obtained from $\mathcal{M}$ using the VMTK~\cite{izzo2018vascular} centerline extraction workflow, resulting in an ordered
set of points $\{\boldsymbol{c}_k\}_{k=1}^{M}$ forming a polyline inside the lumen. We define a discrete arc-length parameter
\begin{equation}
\begin{aligned}
s_0(1)=0,\qquad s_0(k+1)=s_0(k)+\|\boldsymbol{c}_{k+1}-\boldsymbol{c}_k\|,\\
k=1,\dots,M-1,
\end{aligned}
\end{equation}
which is used to parameterize the centerline by arc length and to interpolate quantities at arbitrary $s\in[0,L]$.

\subsubsection{Selection of arc-length samples.}
We select a set of arc-length locations $\{s_i\}_{i=0}^{N}$ with $0=s_0 < s_1 < \dots < s_N=L$. These samples may be
uniform (constant $\Delta s$) or non-uniform (variable $\Delta s_i=s_{i+1}-s_i$), for example by allocating denser
samples in regions of larger centerline curvature and/or in regions where the lumen cross-section changes rapidly.
At each $s_i$, we obtain the corresponding centerline position $\boldsymbol{r}_i=\boldsymbol{r}(s_i)$ and a unit tangent direction $\boldsymbol{t}_i$
estimated from the local centerline geometry.

\subsubsection{Normal-plane slicing.}
At each sample $s_i$, we define the normal plane orthogonal to the tangent,
\begin{equation}
\Pi_i=\left\{\boldsymbol{x}\in\mathbb{R}^3\ \big|\ (\boldsymbol{x}-\boldsymbol{r}_i)^\top \boldsymbol{t}_i = 0\right\},
\end{equation}
and compute its intersection with the surface mesh $\mathcal{M}$. This yields a set of cross-sectional contour
points $\mathcal{C}_i=\{\boldsymbol{p}_{ij}\}_{j=1}^{n_i}$. Since mesh-plane intersection may produce multiple disconnected
polylines due to local irregularities or complex anatomy, we retain the contour component most consistent with the
local lumen section (e.g., the component closest to $\boldsymbol{r}_i$ and/or with the largest extent), and discard spurious
components.

\subsubsection{PCA-based major axis and effective diameter.}
The extracted contours are generally non-elliptic and may contain noise. To robustly estimate an in-plane dominant
direction and an effective diameter, we project the contour points onto $\Pi_i$ and apply principal component
analysis (PCA) \cite{jolliffe2016principal}. We build an orthonormal basis $\{\boldsymbol{e}_{2i},\boldsymbol{e}_{3i}\}$ spanning $\Pi_i$ (with $\boldsymbol{e}_{2i}\perp \boldsymbol{t}_i$ and
$\boldsymbol{e}_{3i}=\boldsymbol{t}_i\times \boldsymbol{e}_{2i}$), and compute 2D coordinates
\begin{equation}
\boldsymbol{q}_{ij}=\Big((\boldsymbol{p}_{ij}-\boldsymbol{r}_i)^\top \boldsymbol{e}_{2i},\ (\boldsymbol{p}_{ij}-\boldsymbol{r}_i)^\top \boldsymbol{e}_{3i}\Big)\in\mathbb{R}^2.
\end{equation}
Let $\boldsymbol{u}_{1i}\in\mathbb{R}^2$ be the first principal direction of $\{\boldsymbol{q}_{ij}\}$, i.e., the largest-variance axis. The endpoints
along this axis are obtained by extremal projection:
\begin{equation}
j^-=\arg\min_j \big(\boldsymbol{q}_{ij}^\top \boldsymbol{u}_{1i}\big),\qquad
j^+=\arg\max_j \big(\boldsymbol{q}_{ij}^\top \boldsymbol{u}_{1i}\big),
\end{equation}
which define the 3D endpoints $\boldsymbol{p}_i^-:=\boldsymbol{p}_{ij^-}$ and $\boldsymbol{p}_i^+:=\boldsymbol{p}_{ij^+}$, as shown in Fig.~\ref{fig:scanmodel} (b). We then compute the effective diameter as
\begin{equation}
a_i=\tfrac{1}{2}\|\boldsymbol{p}_i^+-\boldsymbol{p}_i^-\|.
\end{equation}
Optionally, the PCA eigenvalue ratio can be used as a simple anisotropy indicator to assess the reliability of the
in-plane direction estimate, which becomes ill-conditioned for near-circular sections.

The resulting set of samples $\{s_i,\boldsymbol{r}_i,\boldsymbol{t}_i,a_i,\boldsymbol{p}_i^\pm\}$ provides a compact geometric description of the ST lumen.
In the next section, we use these quantities to construct a discrete local frame along the centerline and enforce
frame continuity prior to building an analytic Cosserat-rod parametrization.
\begin{figure}[t]
	\centering
	\includegraphics[width=0.45\textwidth]{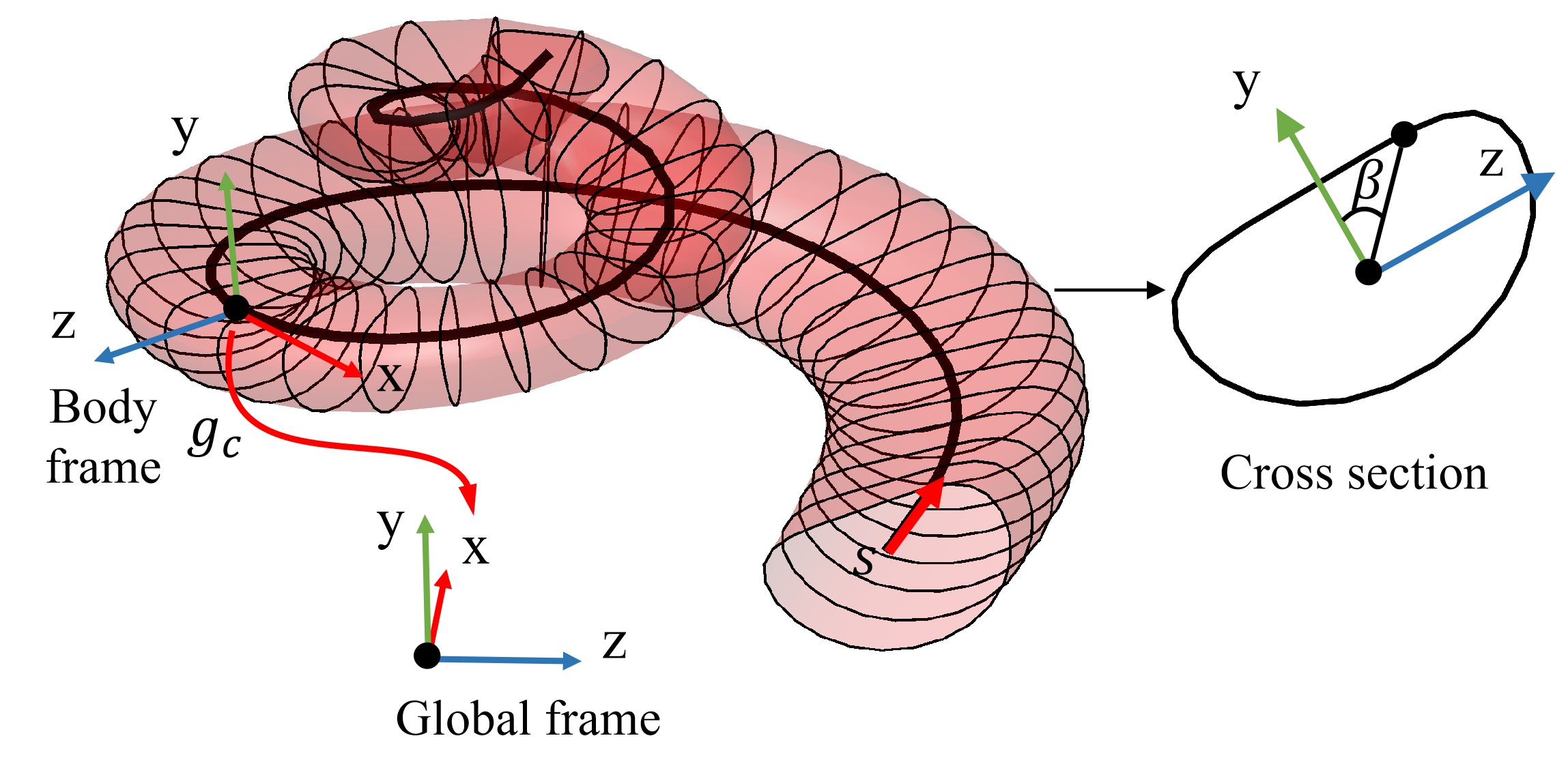}
	\caption{Parametric cochlear lumen model. Cross sections defined by $\boldsymbol{f}_c(s,\beta)$ are swept along the centerline pose curve $\boldsymbol{g}_c(s)\in SE(3)$ to form a continuous, differentiable lumen surface.}
	\label{fig:cochleamodel5}
\end{figure}
\subsection{Discrete configurations at the sampling stations}\label{sec:discrete_configurations}

Section~\ref{sec:centerline_and_sections} provides a set of arc-length samples
$\{s_i,\boldsymbol{r}_i,\boldsymbol{t}_i,a_i,\boldsymbol{p}_i^\pm\}_{i=0}^{N}$ along the scala-tympani centerline, where $\boldsymbol{r}_i\in\mathbb{R}^3$ is the
centerline position, $\boldsymbol{t}_i$ is the local tangent direction, and $\boldsymbol{p}_i^\pm$ are the PCA-based endpoints of the
cross-section contour at station $i$.

At each sampling station, we directly construct a configuration matrix $\boldsymbol{g}_{c,i}\in SE(3)$ by defining a right-handed
local frame whose $x$-axis coincides with the centerline tangent. Specifically, we set
\begin{equation}
\boldsymbol{x}_i := \frac{\boldsymbol{t}_i}{\|\boldsymbol{t}_i\|},
\quad
\boldsymbol{z}_i := \frac{\boldsymbol{p}_i^+ - \boldsymbol{p}_i^-}{\|\boldsymbol{p}_i^+ - \boldsymbol{p}_i^-\|},
\quad
\boldsymbol{y}_i := \frac{\boldsymbol{z}_i \times \boldsymbol{x}_i}{\|\boldsymbol{z}_i \times \boldsymbol{x}_i\|},
\end{equation}
and assemble $\boldsymbol{R}_{c,i}=[\boldsymbol{x}_i\ \boldsymbol{y}_i\ \boldsymbol{z}_i]\in SO(3)$. The discrete pose at station $i$ is then
\begin{equation}
\boldsymbol{g}_{c,i}=
\begin{bmatrix}
\boldsymbol{R}_{c,i} & \boldsymbol{r}_i\\
0 & 1
\end{bmatrix}\in SE(3),\qquad i=0,\dots,N.
\end{equation}
As shown in Fig.~\ref{fig:scanmodel} (c), the resulting sequence $\{\boldsymbol{g}_{c,i}\}$ provides a compact, structured geometric description of the scala tympani, and will
serve as the direct input to the analytic Cosserat-rod parametrization and surface construction presented in the next
section.

\subsection{Analytic geometric exact parametrization and surface construction}\label{sec:analytic_cosserat}

The discrete configurations $\{s_i,\boldsymbol{g}_{c,i}\}_{i=0}^{N}$ constructed in Sec.~\ref{sec:discrete_configurations} define a
compact, structured description of the scala tympani. In this section, we convert these samples into an analytic
geometrically exact representation by assuming piecewise-constant strain between consecutive stations, as shown in Fig.~\ref{fig:scanmodel} (d) and Fig.~\ref{fig:cochleamodel5}. This yields a
closed-form expression of the pose $\boldsymbol{g}_c(s)\in SE(3)$ for any $s\in[0,L]$, and enables continuous evaluation of the
parametrized lumen surface, as shown in Fig.~\ref{fig:scanmodel} (d).

\subsubsection{Piecewise-constant strain.}
Let $\Delta s_i := s_{i+1}-s_i$ denote the arc-length spacing. For each segment
$[s_i,s_{i+1}]$, we define the relative motion
\begin{equation}
\Delta \boldsymbol{g}_{c,i} := \boldsymbol{g}_{c,i}^{-1}\boldsymbol{g}_{c,i+1}\in SE(3),
\end{equation}
and compute a constant body strain (twist) $\boldsymbol{\xi}_{c,i}\in\mathbb{R}^6$ from the $SE(3)$ logarithm,
\begin{equation}
\boldsymbol{\xi}_{c,i} := \frac{1}{\Delta s_i}\, \log\!\big(\Delta \boldsymbol{g}_{c,i}\big)^\vee,
\qquad i=0,\dots,N-1,
\end{equation}
The pose along the
segment is then given analytically by
\begin{equation}
\boldsymbol{g}_c(s)=\boldsymbol{g}_{c,i}\,\exp\!\big(\widehat{\boldsymbol{\xi}}_{c,i}\,(s-s_i)\big),
\qquad s\in[s_i,s_{i+1}],
\end{equation}
with $\widehat{(\cdot)}$ the hat operator and $\exp(\cdot)$ the matrix exponential on $SE(3)$. Although the model is
parameterized by $N$ constant-strain segments, geometric queries (e.g., surface evaluation and collision/contact
checks) can be performed by sampling the analytic expression within each segment at an arbitrary resolution, without
increasing the model dimension.

\subsubsection{Cross-section parameterization along arc length.}
From Sec.~\ref{sec:centerline_and_sections}, each station provides an effective cross-section scale
$a_i$. We define a continuous scale function $a(s)$ by interpolation along arc length, and similarly for any additional cross-section parameters when needed.

Let $\boldsymbol{g}_c(s)=(\boldsymbol{R}_c(s),\boldsymbol{r}(s))$ denote the pose curve obtained above. As illustrated in Fig.~\ref{fig:cochleamodel5}, the scala-tympani lumen surface is defined as a
two-parameter surface indexed by $(s,\beta)$, where $s\in[0,L]$ is the arc-length coordinate and
$\beta\in[0,2\pi)$ is a circumferential parameter around the local tangential axis. We introduce a cross-sectional
contour function $\boldsymbol{f}(s,\beta)\in\mathbb{R}^3$ expressed in the local frame (with $\boldsymbol{f}$ lying in the plane normal to the
rod axis), and map it to the global frame via
\begin{equation}
\boldsymbol{p}_c(s,\beta) = \boldsymbol{r}(s) + \boldsymbol{R}_c(s)\,\boldsymbol{f}(s,\beta), \quad (s,\beta)\in[0,L]\times[0,2\pi).
\end{equation}

In our implementation, in order to approximate the cross section of scala-tympani, the parameterized cross-section is modeled as a vertically asymmetric ellipse (flatter in the upper
half and closer to circular in the lower half), parameterized by a single angle $\beta$. We set
\begin{equation}
\boldsymbol{f}(s,\beta)=
\begin{bmatrix}
0\\
a(s)\cos\beta\\
b(s,\beta)\sin\beta
\end{bmatrix},
\end{equation}
where $a(s)>0$ is the horizontal semi-axis and $b(s,\beta)>0$ is an angle-dependent vertical semi-axis. To obtain a
smooth transition between an ``upper'' semi-axis $b_{\mathrm{u}}(s)$ and a ``lower'' semi-axis $b_{\mathrm{l}}(s)$,
we define the blending weight
\begin{equation}
w(\beta)=\frac{1+\sin\beta}{2}\in[0,1],
\end{equation}
and set
\begin{equation}
b(s,\beta)= b_{\mathrm{l}}(s) - \big(b_{\mathrm{l}}(s)-b_{\mathrm{u}}(s)\big)\,w(\beta)^{\,p},
\quad p\ge 1.
\end{equation}
This construction yields $b(s,\frac{\pi}{2})=b_{\mathrm{u}}(s)$ (top) and $b(s,\frac{3\pi}{2})=b_{\mathrm{l}}(s)$
(bottom). Moreover, because $\sin\beta=0$ at $\beta\in\{0,\pi\}$, the curve is closed and position-continuous, and
the tangent direction is also continuous at $\beta=0,\pi$ (the derivative terms involving $b'(s,\beta)$ vanish when
$\sin\beta=0$). The exponent $p$ controls how localized the ``flattening'' is near the top.

The scalar functions $a(s)$, $b_{\mathrm{u}}(s)$ and $b_{\mathrm{l}}(s)$ are specified along arc length by
interpolating their discrete estimates at $\{s_i\}$ in Sec.~\ref{sec:centerline_and_sections}.

Thus far, we have established an analytic representation of the patient-specific scala-tympani lumen that supports
the geometric quantities required by contact. The next step is to construct candidate contact pairs between the EA and
the lumen surface, i.e., to identify points on the EA and their corresponding closest points on the lumen, together
with the associated normals and gap functions. This contact-pair construction is presented in the next section.

\section{Contact Dynamics}\label{section:dynamics}
This section analyzes the interaction between the cochlear-implant electrode array and the cochlea.

\begin{figure}[t]
	\centering
	\includegraphics[width=0.35\textwidth]{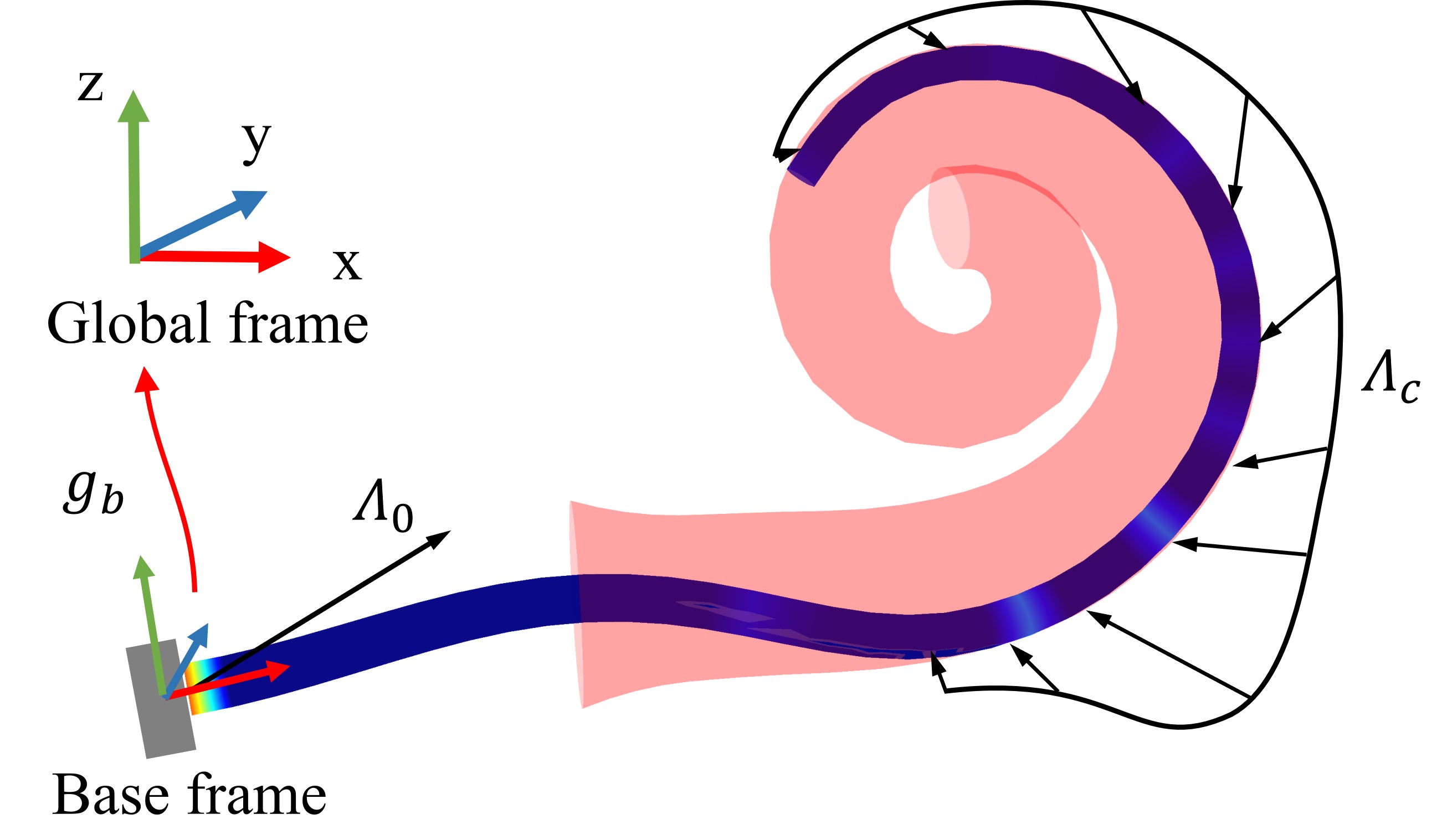}
	\caption{During implantation, an insertion wrench $\boldsymbol{\Lambda}_0$ is applied at the proximal end of implant, while a contact load $\boldsymbol{\Lambda}_c$ acts over the contact region.}
	\label{fig:2rodc2}
\end{figure}

As depicted in Fig.~\ref{fig:2rodc2}, during insertion the proximal end of the implant is driven forward by a support tool. The interaction with the cochlea consists of an externally applied insertion wrench at the proximal end, denoted by $\boldsymbol{\Lambda}_0$, and contact loads generated by the array--cochlea contact. The latter is modeled as a distributed contact load $\boldsymbol{\Lambda}_c$ along the array (including the contribution at the distal end when contact occurs). Together, these actions define the boundary conditions and constraints of the contact dynamics.

We begin by introducing the contact pairs and the procedure used to identify them, since the contact constraints are enforced on these pairs.
\subsection{Contact pair search}\label{section:collision}
\begin{figure}[t]
	\centering
	\includegraphics[width=0.35\textwidth]{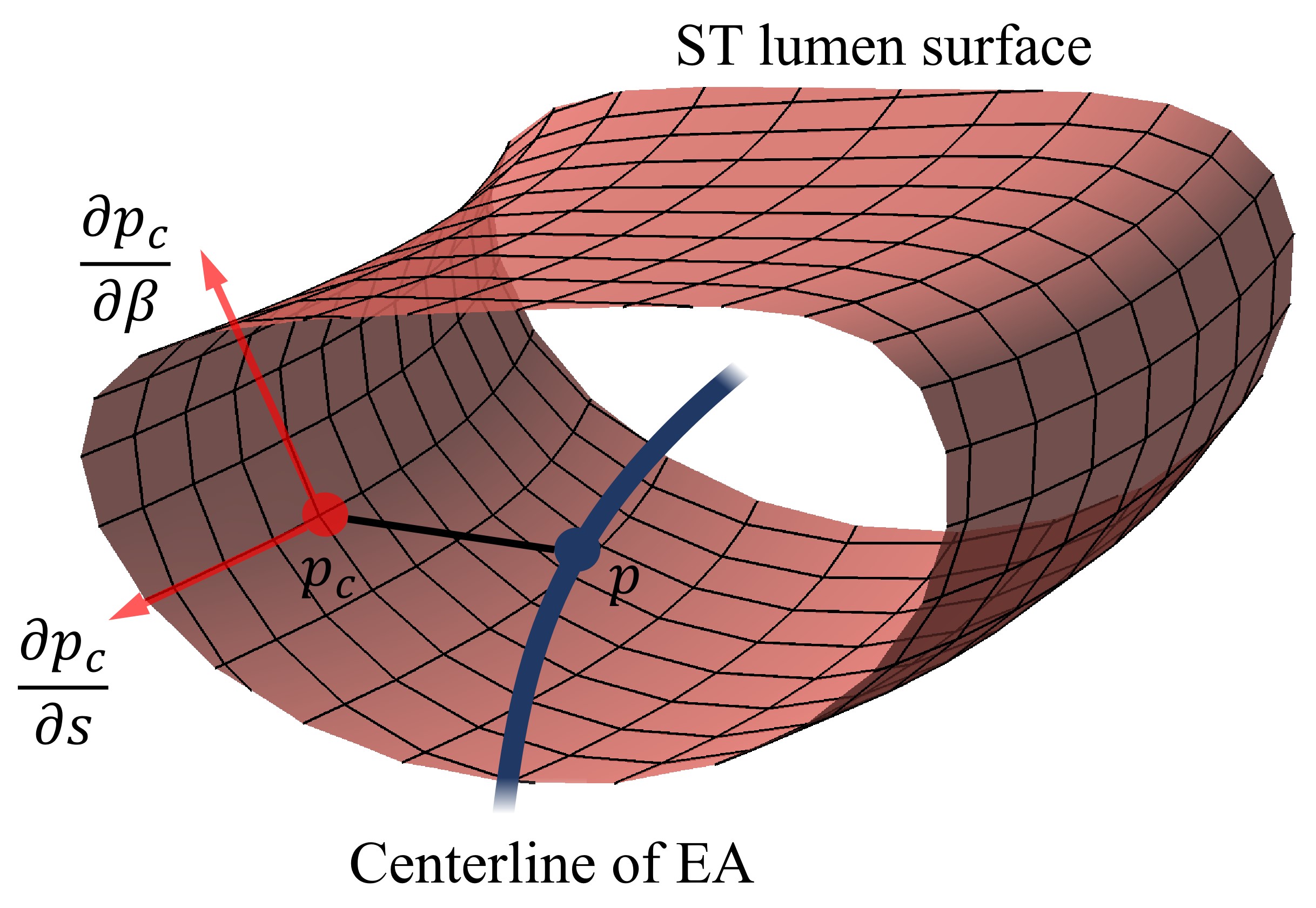}
\caption{Contact-pair search on the CT-derived scala tympani (ST) lumen surface. The lumen is represented by a smooth parametric surface $\boldsymbol{p}_c(s,\beta)$. Given a point $\boldsymbol{p}$ on the EA centerline (blue), the corresponding surface point $\boldsymbol{p}_c$ (red) is obtained by minimizing the Euclidean distance $\|\boldsymbol{p}-\boldsymbol{p}_c(s,\beta)\|$. The surface derivatives $\partial\boldsymbol{p}_c/\partial s$ and $\partial\boldsymbol{p}_c/\partial\beta$ provide the gradients required by this optimization and define the local tangent plane for subsequent contact-frame construction.}
	\label{fig:contactpairsearch}
\end{figure}

As illustrated in Fig.~\ref{fig:contactpairsearch}, a contact pair consists of two points: a {primary} point $\boldsymbol{p}$ on the implant (EA) centerline
and its corresponding {secondary} point $\boldsymbol{p}_c$ on the cochlear lumen surface, defined as the closest
point to $\boldsymbol{p}$ on the parametrized lumen geometry. In practice, we introduce a set of contact stations
$\{\boldsymbol{p}_k\}$ uniformly distributed along the implant centerline, and perform the closest-point query for each
station independently.


Recall from Sec.~\ref{section:geometric} that the cochlear lumen surface admits a smooth two-parameter
parametrization $\boldsymbol{p}_c(\boldsymbol{X})$, with $\boldsymbol{X}=(s,\beta)\in\Omega$ (in our case,
$\Omega=[0,L]\times[0,2\pi)$). As illustrated in Fig.~\ref{fig:contactpairsearch}, for a given primary point $\boldsymbol{p}$, we define its secondary point as
\begin{equation}\label{eq:closest_point_surface}
\boldsymbol{X}
=
\arg\min_{\boldsymbol{X}\in\Omega}\ 
\mathscr{D}(\boldsymbol{X};\boldsymbol{p}),
\end{equation}
with 
\begin{equation}\label{eq:closest_point_surface2}
\mathscr{D}(\boldsymbol{X};\boldsymbol{p})
=
\frac{1}{2}\left\|\boldsymbol{p}_c(\boldsymbol{X})-\boldsymbol{p}\right\|.
\end{equation}

The gradient of
$\mathscr{D}$ with respect to the surface parameters admits the closed form
\begin{equation}\label{eq:closest_point_gradient}
\nabla_{\boldsymbol{X}}\mathscr{D}(\boldsymbol{X};\boldsymbol{p})
=
\frac{\partial \boldsymbol{p}_c(\boldsymbol{X})}{\partial \boldsymbol{X}}^\top(\boldsymbol{p}_c(\boldsymbol{X})-\boldsymbol{p}),
\end{equation}
Since $\boldsymbol{p}_c(\boldsymbol{X})$ is
available analytically from Sec.~\ref{section:geometric}, this gradient can
be evaluated in closed form, enabling efficient gradient-based iterations for the closest-point query.

Recall that the cochlear lumen surface is parametrized by $\boldsymbol{p}_c(\boldsymbol{X})$ with
$\boldsymbol{X}=(s,\beta)\in[0,L]\times[0,2\pi)$, and admits the explicit form
$
\boldsymbol{p}_c(s,\beta)=\boldsymbol{r}(s)+\boldsymbol{R}_c(s)\,\boldsymbol{f}(s,\beta).
$
Within each constant-strain segment of the analytic Cosserat-rod model, the pose satisfies $\boldsymbol{g}_c'(s)=\boldsymbol{g}_c(s)\widehat{\boldsymbol{\xi}}_e$
with constant body strain $\boldsymbol{\xi}_c=[\boldsymbol{\kappa}_c^\top \ \boldsymbol{\epsilon}_c^\top]^\top$, which implies
$\boldsymbol{R}_c'(s)=\boldsymbol{R}_c(s)\tilde{\boldsymbol{\kappa}}_c$
 and 
$\boldsymbol{r}'(s)=\boldsymbol{R}_c(s)\,\boldsymbol{\epsilon}_c$.
Therefore, we can deduce:
\begin{equation}\label{eq:Jc_compact}
\frac{\partial \boldsymbol{p}_c(\boldsymbol{X})}{\partial \boldsymbol{X}}
=
\boldsymbol{R}_c
\begin{bmatrix}
\boldsymbol{\epsilon}_c+\tilde{\boldsymbol{\kappa}}_c\boldsymbol{f}+\partial_s\boldsymbol{f}
&  
\partial_\beta\boldsymbol{f}
\end{bmatrix},
\end{equation}
where $\partial_s\boldsymbol{f}$ and $\partial_\beta\boldsymbol{f}$ are available in closed form from the chosen
cross-section function.

The closest-point projection above establishes, for each primary contact station on the implant, a well-defined
secondary point on the continuous, differentiable cochlear lumen surface. This provides the geometric primitives
required to evaluate the normal gap and tangential directions, and to assemble the subsequent contact constraints and
contact forces. In the next subsection, we formulate these quantities within our contact-mechanics model and derive
the corresponding constraint equations used in the simulation and control framework.

\subsection{Boundary conditions and contact constraints}\label{sec::Bilateral_control_constrain}
\subsubsection{Boundary conditions}
The proximal end (base) of implant at $s=0$ is fixed with the support tool, meaning the initial configuration of the implant aligns with that of the support tool, i.e., $\boldsymbol{g}_0(t)=\boldsymbol{g}_b(t)$,
where $\boldsymbol{g}_0(t)$ denotes the initial configuration tensor of implant at time $t$ and $\boldsymbol{g}_b$ denotes the configuration tensor of the base of support tool at time $t$.
Conversely, the opposite end of the implant is influenced only by contact forces. Therefore, the boundary conditions for this end are contingent upon the occurrence of contact, which will be explored within the following contact constraints.

\subsubsection{No penetration}
\begin{figure}[t]
	\centering
	\includegraphics[width=0.45\textwidth]{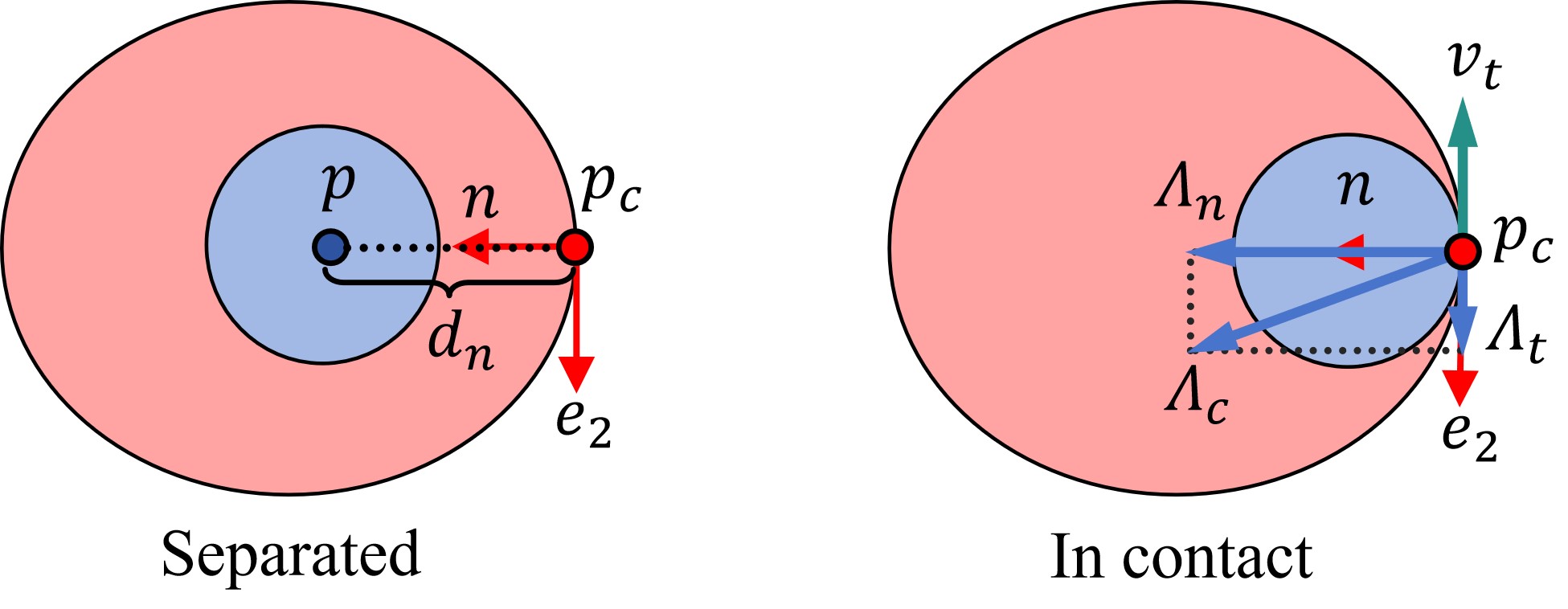}
	\caption{Geometries of the contact pair. The green area represents the cross-section of the implant, while the red area denotes the cross-section of the cochlea.}
	\label{fig:2contact}
\end{figure}

As shown in Fig.~\ref{fig:2contact}, we formulate the contact constraints in a local contact frame defined by three orthonormal unit vectors $(\boldsymbol{n},\boldsymbol{e}_1,\boldsymbol{e}_2)$. The inward unit normal $\boldsymbol{n}$ is perpendicular to the tangent plane of the cochlear wall at the contact point, $\boldsymbol{e}_1$ is aligned with the surface tangent associated with the $s$-direction, and $\boldsymbol{e}_2$ completes the right-handed basis as $\boldsymbol{e}_2=\boldsymbol{n}\times\boldsymbol{e}_1$. Let $\boldsymbol{\tau}_1=\partial \boldsymbol{p}_c(\boldsymbol{X})/\partial s$ and $\boldsymbol{\tau}_2=\partial \boldsymbol{p}_c(\boldsymbol{X})/\partial \beta$ denote two tangent vectors of the cochlear-wall surface. Then,
\begin{equation}
\boldsymbol{n}=\frac{\boldsymbol{\tau}_1\times\boldsymbol{\tau}_2}{\|\boldsymbol{\tau}_1\times\boldsymbol{\tau}_2\|},\qquad
\boldsymbol{e}_1=\frac{\boldsymbol{\tau}_1}{\|\boldsymbol{\tau}_1\|},\qquad
\boldsymbol{e}_2=\boldsymbol{n}\times\boldsymbol{e}_1.
\end{equation}

We define the contact-frame rotation matrix $R_c=[\,\boldsymbol{n}\ \boldsymbol{e}_1\ \boldsymbol{e}_2\,]\in SO(3)$, which maps contact-frame coordinates to the global frame.
Based on this matrix, the configuration matrix $\boldsymbol{g}_c (\boldsymbol{R}_c,\boldsymbol{p}_c)$ can be assembled in order to represent the position and orientation of the contact frame with respect to the global frame.

For each contact pair,
we constrain the normal force and normal gap
the normal contact force for $s\in (0,L]$ through the Signorini's condition \cite{wriggers2006computational}:
\begin{equation}\label{nncp}
	{0}\leq d_n\ \bot\ {\Lambda}_{n}\geq{0}
\end{equation}
where $d_n$ represents the normal distance between the contact pairs $(\boldsymbol{p}, \ \boldsymbol{p}_c)$ expressed in contact frame, i.e.,
\begin{equation}\label{dn}
	d_n=(\boldsymbol{g}_c^{-1}\boldsymbol{g})_{1,4}-r_{EA}
\end{equation}
$(\cdot)_{ij}$ denotes exacting the element in the $i$-th row and $j$-th column of the matrix. Here the first axis of the contact frame is aligned with the inward normal $\boldsymbol{n}$; hence $(g_c^{-1}g)_{1,4}$ is the normal offset expressed in the contact frame.
$r_{EA}$ denotes the radius of implant. 
${\Lambda}_{n}$ represents the \xun{magnitude} of contact force projected on $\boldsymbol{n}$, both should be either positive or zero. The symbol \xunn{``}$\bot$" denotes that the product of $d_n$ and ${\Lambda}_{n}$ equals to zero. This rule establishes a restriction that prevents objects in contact from mutually penetrating, thereby disallowing one object to pass through another.

The introduction of the complementarity constraint in \eqref{nncp} makes the strong form \eqref{contactpde0}
non-smooth due to the $\max(0,x)$ operator. To obtain a differentiable formulation suitable for sensitivity analysis,
we adopt a smooth approximation of the plus function \cite{chen1997smooth}. Specifically, we rewrite the normal gap
$d_n$ and the normal contact force $\Lambda_n$ via a slack variable $u_n$:
\begin{equation}\label{normal}
	d_n = D_\varepsilon(u_n), \qquad \Lambda_n = D_\varepsilon(-u_n),
\end{equation}
where $D_\varepsilon(\cdot)$ is the smooth-plus function
\begin{equation}\label{smooth_plus}
	D_\varepsilon(x)=\frac{1}{2}\left(x+\sqrt{x^2+\varepsilon^2}\right), \qquad \varepsilon>0,
\end{equation}
which satisfies $D_\varepsilon(x)\to \max(0,x)$ as $\varepsilon\to 0$ and is differentiable for all $x$.

\subsubsection{Friction}
\begin{figure}[t]
	\centering
	\includegraphics[width=0.4\textwidth]{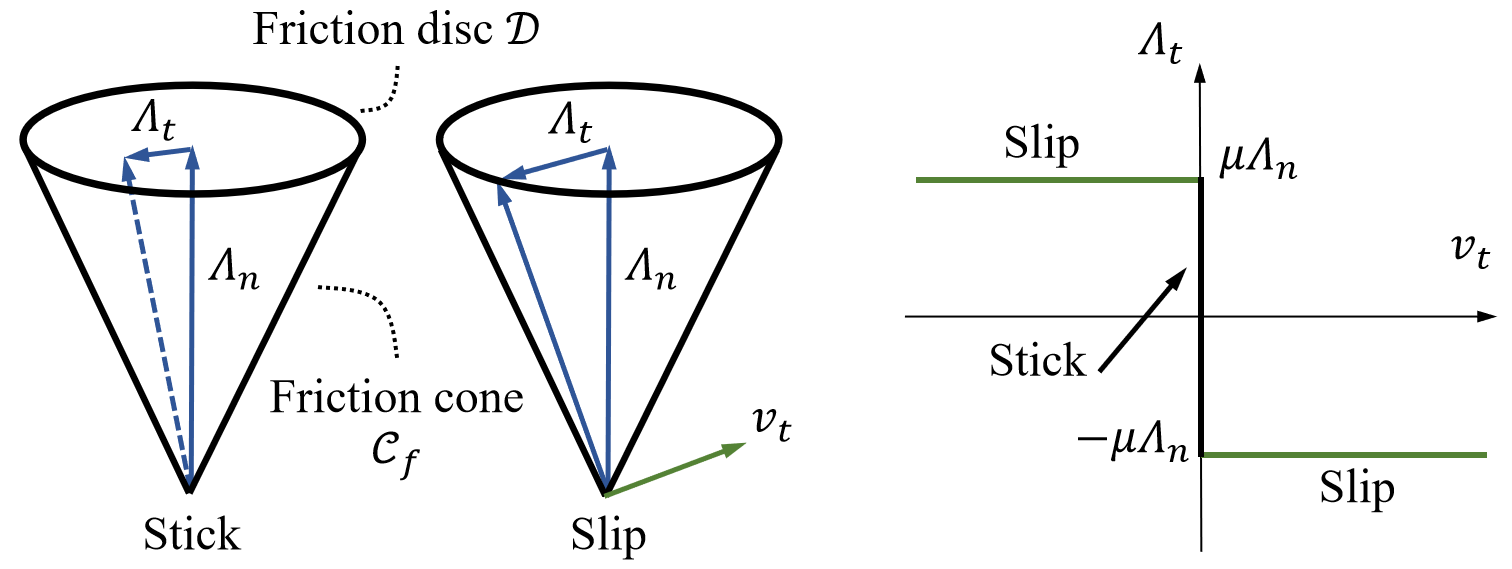}
	\caption{{Coulomb's law of friction states that the reactionary force remains completely within the cone during adhesion between objects, and it aligns with the perimeter of the cone at the onset of slipping.}}
	\label{fig:coulomb}
\end{figure}

We use the Coulomb's law \cite{popov2010contact} to model the friction, {as illustrated in Fig. \ref{fig:coulomb}}. The friction force is constrained within a convex set $\mathcal{C}_f$ corresponding to the friction cone, such that $\boldsymbol{\Lambda}_{t}\in \mathcal{C}_f$.
\xun{The section of the Coulomb's friction cone, i.e., the disk $\mathcal{D}(\mu \Lambda_{n})$ is defined by
\begin{equation}\label{friction cone}
	\mathcal{D}(\mu \Lambda_{n})=\{\boldsymbol{\Lambda}_{t}|\mu \Lambda_{n}-\Vert\boldsymbol{\Lambda}_{t}\Vert\geq0\}
\end{equation}}

The Coulomb's friction cone contains two different states (stick $v_t=0$ or slip $v_t\neq0$) as illustrated in Fig. \ref{fig:coulomb},
where $\mu$ is the coefficient of friction, $\boldsymbol{\Lambda}_t$ is the friction force and $\boldsymbol{v}_t$ is the slip velocity. The Coulomb friction constraints can be expressed as below:
\begin{equation}\label{c222}
	\Vert{\boldsymbol\Lambda}_{t}\Vert\boldsymbol{v}_{t}+ \Vert\boldsymbol{v}_{t}\Vert\boldsymbol{\Lambda}_{t}=\mathbf{0}
\end{equation}
\begin{equation}\label{c333}
	0\leq \Vert\boldsymbol{v}_{t}\Vert \ \bot\ (\mu\Lambda_{n}-\Vert\boldsymbol{\Lambda}_{t}\Vert)\geq0
\end{equation}
Here, equation (\ref{c222}) ensures that the friction force is always opposite to the sliding velocity. (\ref{c333}) ensures the conditions for the occurrence of two types of friction states (stick and slip) and stipulates that only one of these states can occur at any given time. The above two equations can also be reformulated using the plus function, as detailed below:
\begin{equation}\label{friction}
	\boldsymbol{v}_{t}=D(x)\frac{\boldsymbol{u}_t}{\Vert\boldsymbol{u}_t\Vert} \ , \ \boldsymbol{\Lambda}_t=\boldsymbol{v}_t-\boldsymbol{u}_t
\end{equation}
where $x= \Vert\boldsymbol{u}_t\Vert- \mu\Lambda_n$, $\boldsymbol{u}_t$ is the slack variable.

By introducing the slack variable $u_n$ and $\boldsymbol{u}_t$, (\ref{normal}) and (\ref{friction}) establish the contact constraints for one contact pair.

So far, we have \xun{combined} the dynamics equations defined in (\ref{kincos})-(\ref{contactpde0}) with the contact constraints formulated in (\ref{normal}) and (\ref{friction}). It should be noted that these constraints are written w.r.t. contact frame, while the contact force and velocity in Newton-Euler dynamics (\ref{contactpde0}) of our model is written in body frame, thus it is necessary to align the physical quantities, including contact forces and velocities, between two frames. For simplicity, we define the contact force in the contact frame and then transform it into the body frame. Similarly, velocity is defined in the body frame and subsequently transformed into the contact frame. In the following subsection, we will outline the methods used for these transformations.
\subsection{The transfer between body frame and contact frame}
\begin{figure}[t]
	\centering
	\includegraphics[width=0.47\textwidth]{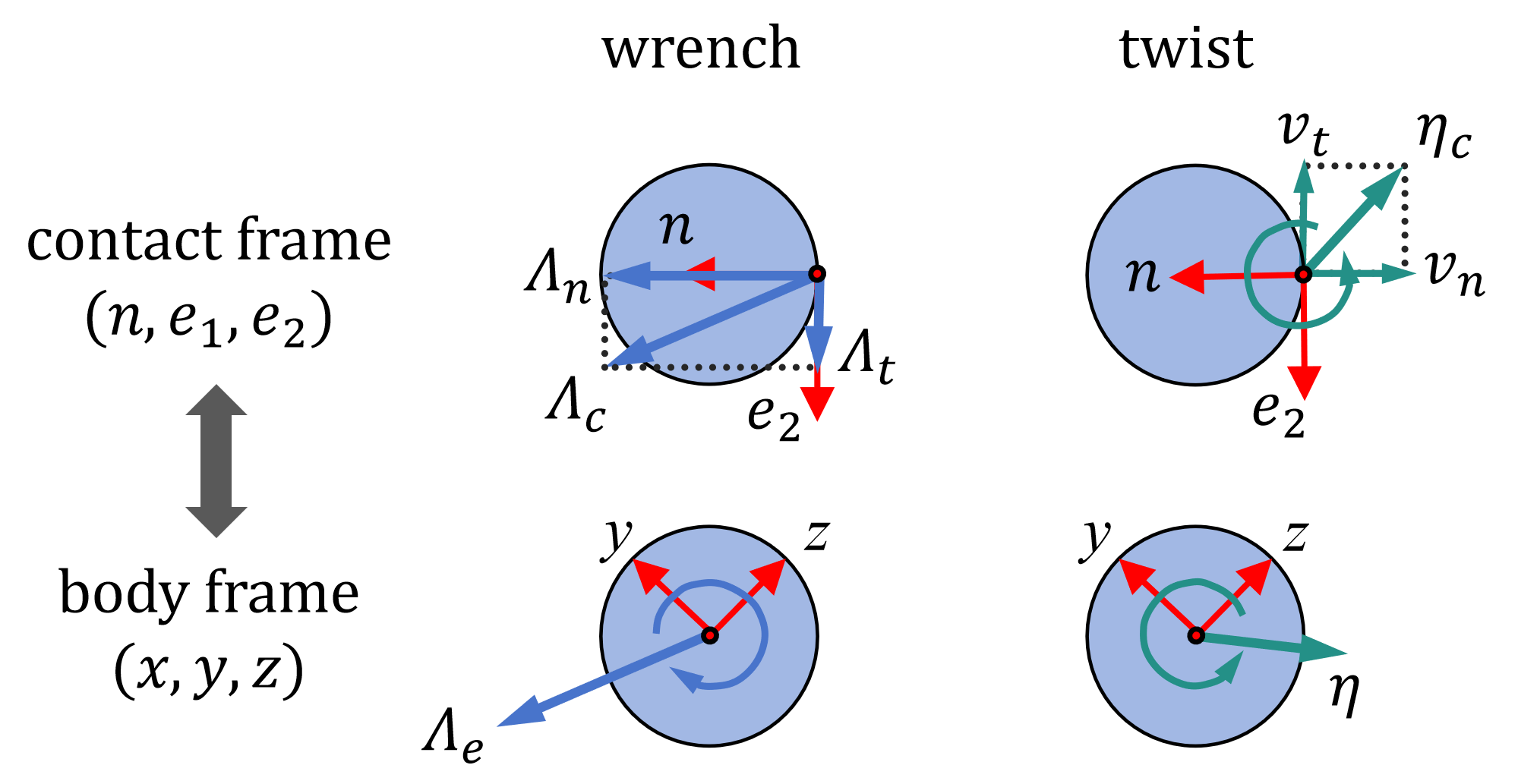}
	\caption{The transformation of contact wrench and velocity twist between the body frame and the contact frame.}
	\label{fig:contact1}
\end{figure}
Defining $\boldsymbol{\Lambda}_c=[
	\boldsymbol{0}_{1\times3}\ {\Lambda}_n \ \boldsymbol{\Lambda}^\top_t]^\top\in\mathbb{R}^6$ as the contact force written in contact frame, since the Cosserat rod dynamics are formulated within the body frame and all external forces act along the rod's centerline, integrating contact \xun{wrench} $\boldsymbol{\Lambda}_c$ into the cochlear implant's dynamics requires specific adjustments. Contact wrench $\boldsymbol{\Lambda}_c$ at the points of contact must be translated to the mass center of their respective ``disc'' to calculate the equivalent external wrench, denoted as $\boldsymbol{\Lambda}_e$. These wrenches are then \xun{expressed in} the body frame, as depicted in Fig. \ref{fig:contact1}.
According to screw dynamics \cite{selig2004lie}, 
the transfer from $\boldsymbol{\Lambda}_c$ to $\boldsymbol{\Lambda}_e$ in body frame can be achieved by the following formula:   
\begin{equation}\label{trf}
		\boldsymbol{\Lambda}_e=\mathrm{Ad}^\top_{\boldsymbol{g}_c^{-1}\boldsymbol{g}}{\boldsymbol{\Lambda}}_c
\end{equation}
where $\mathrm{Ad}_{(\cdot)}$ is transformation matrix defined in Appendix \ref{notations2}.
Recalling that the dot product of a force and a velocity is a power which is a coordinate-independent quantity, therefore, we have
\begin{equation}
	\boldsymbol{\Lambda}_e^{\top}\boldsymbol{\eta}=\boldsymbol{\Lambda}_c^{\top}\boldsymbol{\eta}_c
	\label{power}
\end{equation}
where $\boldsymbol{\eta}$ denotes the velocity twist of mass center expressed in body frame and $\boldsymbol{\eta}_c$ denotes the velocity twist of contact point expressed in the contact frame. Substituting (\ref{trf}) into (\ref{power}) yields
$\boldsymbol{\Lambda}_e^{\top}\boldsymbol{\eta}_c=(\mathrm{Ad}^\top_{\boldsymbol{g}^{-1}\boldsymbol{g}_{c}}{\boldsymbol{\Lambda}}_c)^{\top}\boldsymbol{\eta}$
which always holds for all $\boldsymbol{\Lambda}_c^{\top}$, and simplifies to
\begin{equation}		\boldsymbol{\eta}_c=\mathrm{Ad}_{\boldsymbol{g}_c^{-1}\boldsymbol{g}}\boldsymbol{\eta}.
\end{equation}
Subsequently, the sliding velocity of the contact point expressed in the contact frame is:
\begin{equation}\label{vteta}
		\boldsymbol{v}_t=(\mathrm{Ad}_{\boldsymbol{g}_c^{-1}\boldsymbol{g}}{\boldsymbol{\eta}})_{5:6}
\end{equation}
where $(\cdot)_{i:j}$ denotes the subvector formed by taking entries $i$ through $j$.

Thus far, we can summarize the contact constraint imposed on the implant. For a contact pair $(\boldsymbol{p},\boldsymbol{p}_c)$, we introduce the slack variable $\boldsymbol{u}=[u_n \ \ \boldsymbol{u}_t^\top]^\top$. The contact wrench applied to the implant is then constrained as
\begin{equation}
    \boldsymbol{\Lambda}_e(\boldsymbol{g},\boldsymbol{u}) = \mathrm{Ad}^\top_{\boldsymbol{g}_c^{-1}\boldsymbol{g}} \begin{bmatrix}
        D(-u_n) \\ D(x) \boldsymbol{t}-\boldsymbol{u}_t \\ \boldsymbol{0}_{3\times 1}
    \end{bmatrix}
\end{equation}
with $x=\Vert\boldsymbol{u}_t\Vert- \mu D(-u_n)$ and $\boldsymbol{t}=\boldsymbol{u}_t/\Vert \boldsymbol{u}_t \Vert_2$.
The position and velocity are constrained by the implicit function:
\begin{equation}
\boldsymbol{f}_c(\boldsymbol{g},\boldsymbol{\eta},\boldsymbol{u})=
\begin{bmatrix}
    (\boldsymbol{g}_c^{-1}\boldsymbol{g})_{1,4}-r - D(u_n)\\
    (\mathrm{Ad}_{\boldsymbol{g}_c^{-1}\boldsymbol{g}}{\boldsymbol{\eta}})_{5:6}-D(x) \boldsymbol{t}
\end{bmatrix}=\boldsymbol{0}.
\end{equation}
\subsection{Contact pseudo-dynamics of implant}
Cochlear implant insertion is performed slowly, and the implant remains close to quasi-static equilibrium
throughout the procedure. Therefore, inertial effects are negligible and we do not include mass and acceleration
terms in the governing equations. However, a purely static formulation is still inadequate in the presence of
frictional contact, because friction depends on the relative sliding velocity. In particular, when the contact
state transitions from stick to slip, a strictly static model may yield ill-posed updates and can
lead to large, discontinuous jumps in the computed configuration.

To obtain a well-posed time evolution and a continuous trajectory, we introduce a pseudo-dynamics model by adding a
small viscous regularization to the internal force. This weak damping provides temporal smoothing without
significantly altering the quasi-static nature of the insertion, while enabling robust handling of stick--slip
transitions in frictional contact.

After defining all constraints and implementing the transformation of contact forces and velocities across different
reference frames, we can now formalize the strong form of the dynamics for the cochlear implant under contact
constraints. 
\subsubsection{Strong form of contact pseudo-dynamics}
Introducing the internal force $\boldsymbol{\Lambda}_{i}=\boldsymbol{\mathcal{K}}{\boldsymbol{\xi}}+\boldsymbol{\mathcal{D}}\dot{\boldsymbol{\xi}}$, where $\mathcal{K}$ and $\mathcal{D}$ denote the stiffness and viscous matrix respectively (readers are referred to \cite{8500341} for detailed definition), and ignoring the inertial force, the strong form of pseudo-dynamics with contact constraints can be rewritten as a set of partial differential equations.
\begin{equation}\label{contactpde}
	\begin{aligned}	\boldsymbol{\Lambda}_{i}^{\prime}-\operatorname{ad}_{\boldsymbol{\xi}}^{\top} \boldsymbol{\Lambda}_{i}+{\boldsymbol{\Lambda}}_e(\boldsymbol{g},\boldsymbol{u}) +{\boldsymbol{\Lambda}}_0 = 0
	\end{aligned}
\end{equation}
satisfying the following boundary condition:
\begin{equation}\label{contactpdec}
	\boldsymbol{g}_0 = \boldsymbol{g}_b
\end{equation}
as well as the contact constraints:
\begin{equation}\label{constraints2}	 \boldsymbol{f}_c(\boldsymbol{g},\boldsymbol{\eta},\boldsymbol{u})= \boldsymbol{0}
\end{equation}

\subsubsection{Weak form of contact dynamics}
Equations~\eqref{contactpde}--\eqref{constraints2} define a strong Cosserat-rod model with contact.
To obtain a finite-dimensional model suitable for simulation and optimization, we adopt a weak discretization in space and derive a compact DAE system \cite{10494907}.

Following the piecewise-linear strain (PLS) approach \cite{10027557}, we parameterize the strain field by generalized coordinates $\boldsymbol{q}\in\mathbb{R}^{N}$,
\begin{equation}
\boldsymbol{\xi}(s,t)=\boldsymbol{\xi}_0(s)+\boldsymbol{\Phi}_{\xi}(s)\boldsymbol{q}(t),
\quad
\boldsymbol{\eta}(s,t)=\boldsymbol{J}(s,t)\dot{\boldsymbol{q}}(t),
\end{equation}
where $\boldsymbol{\Phi}_{\xi}(s)$ denotes the $C^0$ piecewise-linear basis functions and $\boldsymbol{J}(s,t)$ is the kinematic Jacobian \cite{10027557}.
Using the same type of piecewise-linear basis functions, we discretize the contact slack field as
\begin{equation}
\boldsymbol{u}(s,t)=\boldsymbol{\Phi}_{u}(s)\boldsymbol{\lambda}(t),
\qquad \boldsymbol{\lambda}\in\mathbb{R}^{M},
\end{equation}
where $\boldsymbol{\Phi}_{u}(s)$ is a $C^0$ piecewise-linear basis in the contact space.

Substituting these approximations into the strong form and applying Galerkin projection yields the weak forms
\begin{equation}\label{eq:dynamic_weak1}
\int_{0}^{L}\boldsymbol{J}^\top
\big(\boldsymbol{\Lambda}_{i}^{\prime}-\operatorname{ad}_{\boldsymbol{\xi}}^{\top}\boldsymbol{\Lambda}_{i}
+\boldsymbol{\Lambda}_{e}(\boldsymbol{g},\boldsymbol{u})
+\boldsymbol{\Lambda}_{0}\big)\,\mathrm{d}s
= \boldsymbol{0},
\end{equation}
\begin{equation}\label{w11}
\int_{0}^{L}\boldsymbol{\Phi}_{u}^\top\,\boldsymbol{f}_c(\boldsymbol{g},\boldsymbol{\eta},\boldsymbol{u})\,\mathrm{d}s
= \boldsymbol{0}.
\end{equation}
The resulting contact-aware implant model can be written in the compact DAE form
\begin{align}
\boldsymbol{D}\dot{\boldsymbol{q}}+\boldsymbol{K}\boldsymbol{q}+\boldsymbol{F}_e(\boldsymbol{q},\boldsymbol{\lambda})+\boldsymbol{F}_0(\boldsymbol{q},\boldsymbol{\Lambda}_0) &= \boldsymbol{0}, \label{dynamic}\\
\boldsymbol{g}_0-\boldsymbol{g}_b &= \boldsymbol{0}, \label{keyd1a}\\
\boldsymbol{C}(\dot{\boldsymbol{q}},\boldsymbol{q},\boldsymbol{\lambda}) &= \boldsymbol{0}. \label{keyd1b}
\end{align}
All matrices and vectors in \eqref{dynamic}--\eqref{keyd1b} are defined as follows:
\begin{itemize}
    \item[$\bullet$] $\boldsymbol{D}=\int_0^L \boldsymbol{\Phi}_{\xi}^\top\mathcal{D}\boldsymbol{\Phi}_{\xi}\,\mathrm{d}s\in \mathbb{R}^{N\times N}$, the viscous matrix;
    \item[$\bullet$] $\boldsymbol{K}=\int_0^L \boldsymbol{\Phi}_{\xi}^\top\mathcal{K}\boldsymbol{\Phi}_{\xi}\,\mathrm{d}s\in \mathbb{R}^{N\times N}$, the stiffness matrix;
    \item[$\bullet$] $\boldsymbol{F}_e(\boldsymbol{q},\boldsymbol{\lambda})=\int_0^L \boldsymbol{J}^\top\boldsymbol{\Lambda}_e\,\mathrm{d}s\in \mathbb{R}^{N}$, the reduced contact contribution;
    \item[$\bullet$] $\boldsymbol{F}_0(\boldsymbol{q},\boldsymbol{\Lambda}_0)=\boldsymbol{J}_0^\top\boldsymbol{\Lambda}_0\in \mathbb{R}^{N}$, the reduced boundary wrench contribution, where $\boldsymbol{J}_0=\boldsymbol{J}(0,t)$;
    \item[$\bullet$] $\boldsymbol{C}(\dot{\boldsymbol{q}},\boldsymbol{q},\boldsymbol{\lambda})=\int_{0}^{L}\boldsymbol{\Phi}_{\boldsymbol{u}}^\top
	      \boldsymbol{f}_c\,\mathrm{ds}\in \mathbb{R}^{M}$, the  contact constraint.	
\end{itemize}
For simulation, the DAEs are discretized in time using an implicit Euler scheme \cite{negrut2003implicit}. Together with the collision detection in Sec.~\ref{section:collision}, the current state $(\boldsymbol{q},\dot{\boldsymbol{q}},\boldsymbol{\lambda})$ is obtained at each step.

\section{Insertion Path Planning}\label{section:control}
\subsection{RCM-like constraint and admissible base motion}\label{sec:rcm_constraint}
\begin{figure}[t]
	\centering
	\includegraphics[width=0.45\textwidth]{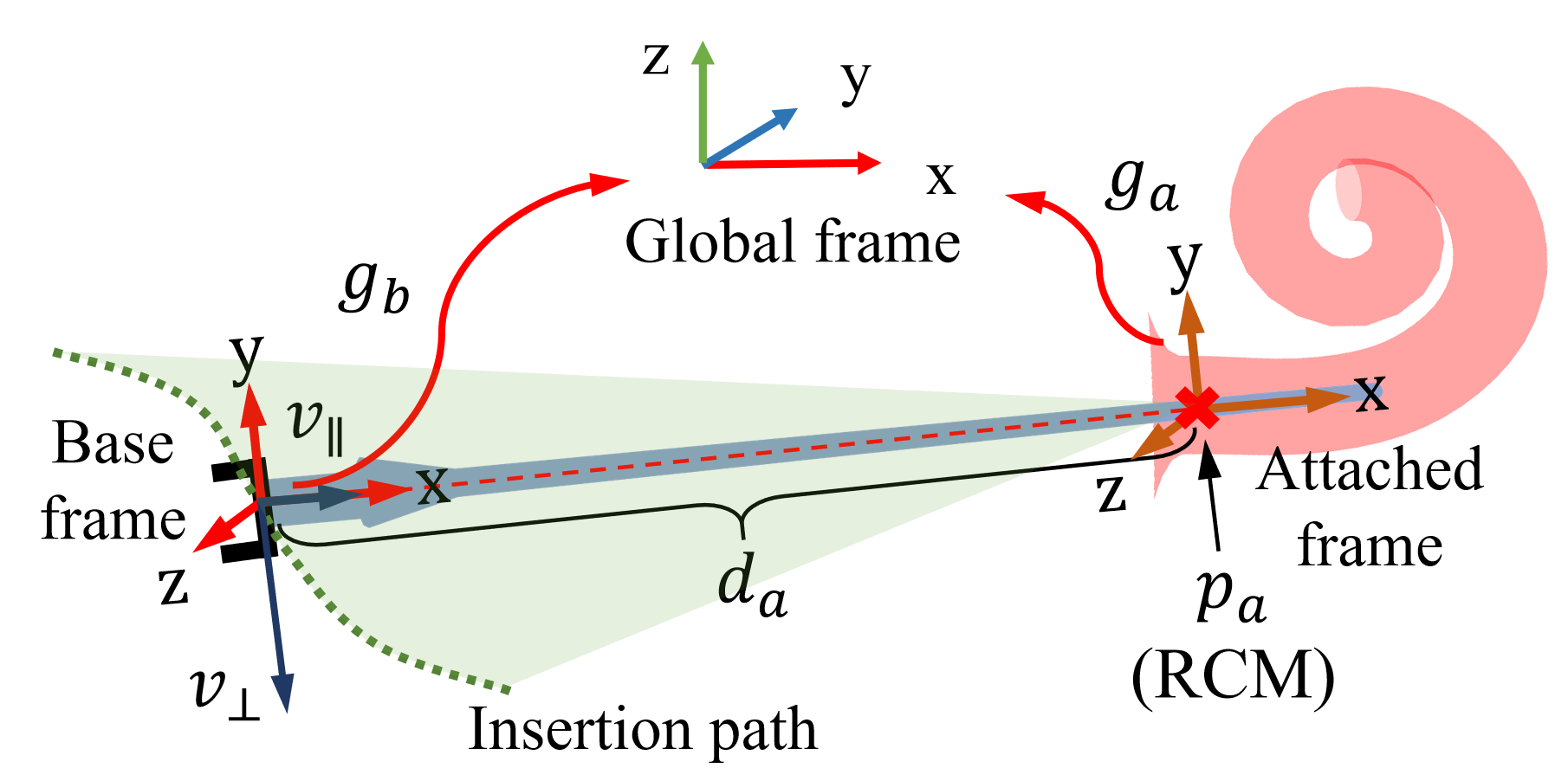}
\caption{RCM-like constraint at the cochlear entrance. The insertion axis is constrained to pass through the virtual RCM point $\boldsymbol{p}_a$. The base-tool velocity is decomposed into a radial component $\boldsymbol{v}_{\parallel}$ (toward $\boldsymbol{p}_a$) and a tangential component $\boldsymbol{v}_{\perp}$ (pivoting about $\boldsymbol{p}_a$).}
	\label{fig:baseimplant2}
\end{figure}

In robotic-assisted minimally invasive surgery, a remote center of motion (RCM) constraint is commonly used to ensure that the tool shaft pivots about a fixed point (e.g., a trocar) while permitting reorientation. In cochlear implantation, although there is no physical trocar, the cochlear entrance provides a natural geometric reference for a similar constraint. We therefore introduce an {RCM-like} constraint by defining the center of the cochlear entrance as a virtual RCM point, denoted by $\boldsymbol{p}_a$ (Fig.~\ref{fig:baseimplant2}). Under this constraint, the implant base (support tool) pivots about $\boldsymbol{p}_a$ and advances toward it.

Let $\boldsymbol{g}_b(t)\in SE(3)$ denote the configuration of the implant base,
\begin{equation}
\boldsymbol{g}_b(t)=
\begin{bmatrix}
\boldsymbol{R}_b(t) & \boldsymbol{p}_b(t)\\
\boldsymbol{0}^\top & 1
\end{bmatrix},
\end{equation}
where $\boldsymbol{R}_b(t)\in SO(3)$ and $\boldsymbol{p}_b(t)\in\mathbb{R}^3$ are its orientation and position, respectively. In free space, $\boldsymbol{g}_b$ has six degrees of freedom. The proposed RCM-like constraint restricts the translational motion to a one-dimensional axial motion while maintaining three rotational degrees of freedom, thereby reducing the admissible motion to $4$ degrees of freedom.

Throughout insertion, the insertion axis of the base (the $x$-axis of the base frame) is required to point toward the virtual RCM point $\boldsymbol{p}_a$. Defining
\begin{equation}
\boldsymbol{a}(t)=\boldsymbol{R}_b(t)\boldsymbol{e}_x,\qquad \boldsymbol{e}_x=[1,0,0]^\top,
\end{equation}
the constraint is enforced by requiring that the vector from the base origin to $\boldsymbol{p}_a$ is collinear with $\boldsymbol{a}(t)$ and oriented along $+\boldsymbol{a}(t)$, i.e.,
\begin{equation}\label{eq:axis_pointing}
\boldsymbol{p}_a-\boldsymbol{p}_b(t)=d_a(t)\,\boldsymbol{a}(t),\qquad d_a(t)\ge 0,
\end{equation}
where $d_a(t)=\|\boldsymbol{p}_a-\boldsymbol{p}_b(t)\|$ is the axial distance from the base to the virtual RCM point. Equation~\eqref{eq:axis_pointing} explicitly states that the base $x$-axis always passes through $\boldsymbol{p}_a$ and points toward it.

For later derivations, it is convenient to rewrite \eqref{eq:axis_pointing} in an equivalent group form. We introduce a frame $\mathcal{A}$ attached at $\boldsymbol{p}_a$ that shares the same orientation as the base:
\begin{equation}
\boldsymbol{g}_a(t)=
\begin{bmatrix}
\boldsymbol{R}_b(t) & \boldsymbol{p}_a\\
\boldsymbol{0}^\top & 1
\end{bmatrix}.
\end{equation}
Then, the base configuration is obtained from $\boldsymbol{g}_a$ by a pure translation of $-d_a(t)$ along the $x$-axis,
\begin{equation}\label{eq:rcm_gb}
\boldsymbol{g}_b(t)=\boldsymbol{g}_a(t)\,
\begin{bmatrix}
1&0&0&-d_a(t)\\
0&1&0&0\\
0&0&1&0\\
0&0&0&1
\end{bmatrix}.
\end{equation}
This representation makes explicit that, once $\boldsymbol{p}_a$ is fixed, the base position $\boldsymbol{p}_b(t)$ is not independent: it is fully determined by $\boldsymbol{R}_b(t)$ and the scalar distance $d_a(t)$ through \eqref{eq:axis_pointing} (equivalently, \eqref{eq:rcm_gb}). Consequently, the admissible motion under the RCM-like constraint consists of a reorientation of $\boldsymbol{R}_b(t)$ (3 DoF) and an axial advancement governed by $d_a(t)$ (1 DoF).

The admissible base velocity can be interpreted as the superposition of (i) a pivoting motion about $\boldsymbol{p}_a$ induced by the base angular velocity (expressed in body frame) $\boldsymbol{\omega}_b(t) = (\boldsymbol{R}_b^\top\dot{\boldsymbol{R}}_b)^\vee$, and (ii) an axial advancement toward $\boldsymbol{p}_a$ with speed $v_{\parallel}(t)=-\dot{d}_a\ge 0$ along the insertion axis. In the following, $v_{\parallel}(t)$ is specified by the insertion protocol, while $\boldsymbol{\omega}_b(t)$ will be adjusted to reduce the contact-induced lateral force at the base.

\subsection{Control input and lateral-force objective}\label{sec:input_objective}
Under the RCM-like constraint in Sec.~\ref{sec:rcm_constraint}, the admissible motion of the implant base consists of a reorientation (three rotational DoFs) and an axial advancement (one translational DoF). In this work, the axial advancement is prescribed by the insertion protocol, while the reorientation is adjusted online to mitigate buckling.

We take the base angular velocity as the control input,
\begin{equation}\label{eq:control_input}
\boldsymbol{u}(t)=\boldsymbol{\omega}_b(t)\in\mathbb{R}^3,
\end{equation}
and treat the axial insertion speed $v_{\parallel}(t)\ge 0$ as a given command (constant in our implementation). The base configuration $\boldsymbol{g}_b(t)$ then evolves according to the kinematics induced by $\boldsymbol{\omega}_b(t)$ together with the RCM-like constraint in \eqref{eq:axis_pointing}--\eqref{eq:rcm_gb}.

As defined in our mechanical model, the base insertion wrench $\boldsymbol{\Lambda}_0(t)\in\mathbb{R}^6$ is expressed in the base (local) frame, whose $x$-axis coincides with the implant axis. We decompose it into a torque part and a force part,
\begin{equation}
\boldsymbol{\Lambda}_0
=
\begin{bmatrix}
\tau_{0x}& \tau_{0y}& \tau_{0z}&
f_{0x}& f_{0y}& f_{0z}
\end{bmatrix}^\top,
\end{equation}
where $\boldsymbol{\tau}_0\in\mathbb{R}^3$ and $\boldsymbol{f}_0\in\mathbb{R}^3$ denote the contact-induced torque and force at the base, respectively. The lateral force components orthogonal to the insertion direction are therefore the $y$- and $z$-components $f_{0y}$ and $f_{0z}$. Since compression-driven buckling is promoted by lateral loading, our objective is to reduce these transverse components while allowing the axial component $f_{0x}$ to transmit the propulsion effort.

We define the instantaneous objective function as
\begin{equation}\label{eq:objective_lateral}
\mathcal{H}(t)
=
\frac{1}{2}\Big(f_{0y}^2(t)+f_{0z}^2(t)\Big).
\end{equation}
In the following, we derive how the base angular velocity $\boldsymbol{\omega}_b(t)$ influences the time evolution of $\boldsymbol{\Lambda}_0(t)$ through the contact-aware implant model, and we design a regulation law to drive $\big[f_{0y},f_{0z}\big]^\top$ toward zero.

\subsection{Differentiated equilibrium--constraint system}\label{sec:diff_system}
In order to derive a continuous-time feedback law for path optimization, we differentiate the quasi-static equilibrium and the kinematic/contact constraints with respect to time.
Throughout this section, all wrenches and twists are expressed in the base (end-effector) body frame.
Moreover, following the implementation-oriented linearization adopted in this work, we treat the contact generalized force $\boldsymbol{F}_e$ and the boundary Jacobian $\boldsymbol{J}_0$ as {lagged} quantities within each control cycle: their geometric dependence on the current configuration is frozen at the linearization point, whereas the normal contact gap still varies with the current configuration.

\subsubsection{Quasi-static equilibrium}
Neglecting inertial and viscous contributions, the weak-form implant model yields the quasi-static force balance
\begin{equation}\label{eq:qs_balance}
\boldsymbol{R}(\boldsymbol{q},\boldsymbol{\lambda},\boldsymbol{\Lambda}_0)
:= \boldsymbol{K}\boldsymbol{q}+\boldsymbol{F}_e(\boldsymbol{\lambda})
+\boldsymbol{J}_0^{\top}\boldsymbol{\Lambda}_0=\boldsymbol{0},
\end{equation}
where $\boldsymbol{q}\in\mathbb{R}^{N}$ denotes the reduced strain coordinates, $\boldsymbol{\lambda}\in\mathbb{R}^{M}$ collects the slack/contact variables, and $\boldsymbol{\Lambda}_0\in\mathbb{R}^{6}$ is the base insertion wrench (Lagrange multiplier) introduced in the contact mechanics model.
Differentiating \eqref{eq:qs_balance} with respect to time gives
\begin{equation}\label{eq:qs_balance_dot}
\boldsymbol{K}\dot{\boldsymbol{q}}
+\boldsymbol{F}_{\lambda}\dot{\boldsymbol{\lambda}}
+\boldsymbol{J}_0^{\top}\dot{\boldsymbol{\Lambda}}_0
=\boldsymbol{0},
\end{equation}
where $\boldsymbol{F}_{\lambda}:=\partial \boldsymbol{F}_e/\partial \boldsymbol{\lambda}\in\mathbb{R}^{N\times M}$ is evaluated at the current linearization point.

\subsubsection{Boundary kinematics under the RCM constraint}
The base configuration $\boldsymbol{g}_b$ is constrained by the RCM-like kinematics at the cochlear entrance: its $x$-axis is maintained to pass through the virtual RCM point $\boldsymbol{p}_a$ and the base advances along its local $x$-axis with speed $v_{\Vert}$, while a body angular velocity $\boldsymbol{\omega}_b$ induces a tangential motion on the RCM sphere.
Accordingly, the base body-frame twist is written as
\begin{equation}\label{eq:etab_omega_v}
\boldsymbol{\eta}_b
=
\begin{bmatrix}
\boldsymbol{\omega}_b\\
\boldsymbol{v}_b
\end{bmatrix}
=
\boldsymbol{G}_{\omega}\boldsymbol{\omega}_b+\boldsymbol{G}_{v}v_{\Vert},
\end{equation}
with
\begin{equation}\label{eq:Gomega_Gv}
\boldsymbol{G}_{\omega}=
\begin{bmatrix}
1&0&0\\
0&1&0\\
0&0&1\\
0&0&0\\
0&0&-d_a\\
0&d_a&0
\end{bmatrix},
\qquad
\boldsymbol{G}_{v}=
\begin{bmatrix}
0\\0\\0\\
1\\0\\0
\end{bmatrix},
\end{equation}
where $d_a$ is the instantaneous distance between the base and the virtual RCM point along the base $x$-axis.
At the implant base, the boundary constraint $\boldsymbol{g}(0,t)=\boldsymbol{g}_b(t)$ implies equality of body twists,
\begin{equation}\label{eq:eta0_etab}
\boldsymbol{\eta}_0=\boldsymbol{\eta}_b,
\end{equation}
and using the reduced kinematics $\boldsymbol{\eta}_0=\boldsymbol{J}_0\dot{\boldsymbol{q}}$ yields the differentiated boundary relation
\begin{equation}\label{eq:bc_dotq}
\boldsymbol{J}_0\dot{\boldsymbol{q}}
=\boldsymbol{G}_{\omega}\boldsymbol{\omega}_b+\boldsymbol{G}_{v}v_{\Vert}.
\end{equation}

\subsubsection{Differentiated contact constraints}
Let $\boldsymbol{C}(\boldsymbol{q},\dot{\boldsymbol{q}},\boldsymbol{\lambda})=\boldsymbol{0}$ denote the assembled equality-form contact constraints introduced in Sec.~\ref{sec::Bilateral_control_constrain}.
In this section we decompose them into a normal part and a tangential (frictional) part and adopt the following linearization:
(i) the normal gap depends on the current configuration and is differentiated with respect to $\boldsymbol{q}$ and $\boldsymbol{\lambda}$;
(ii) the frictional constraint depends on the relative tangential velocity, and the tangential velocity at the current control cycle is supposed to be frozen, such that the frictional constraint is differentiated only with respect to $\boldsymbol{q}$ and $\boldsymbol{\lambda}$.
This yields
\begin{equation}\label{eq:Cn_dot}
\boldsymbol{C}_{q}\dot{\boldsymbol{q}}+\boldsymbol{C}_{\lambda}\dot{\boldsymbol{\lambda}}=\boldsymbol{0},
\end{equation}
where $\boldsymbol{C}_{q}:=\partial \boldsymbol{C}/\partial \boldsymbol{q}$ and $\boldsymbol{C}_{\lambda}:=\partial \boldsymbol{C}/\partial \boldsymbol{\lambda}$ are evaluated at the current linearization point.

\subsubsection{Compact block form}
Collecting \eqref{eq:qs_balance_dot}, \eqref{eq:bc_dotq}, and \eqref{eq:Cn_dot} gives the differentiated equilibrium--constraint system
\begin{equation}\label{eq:block_diff_system}
\underbrace{\begin{bmatrix}
\boldsymbol{K} & \boldsymbol{J}_0^{\top} & \boldsymbol{F}_{\lambda}\\
\boldsymbol{J}_0 & \boldsymbol{0} & \boldsymbol{0}\\
\boldsymbol{C}_{q} & \boldsymbol{0} & \boldsymbol{C}_{\lambda}
\end{bmatrix}}_{\boldsymbol{A}}
\begin{bmatrix}
\dot{\boldsymbol{q}}\\
\dot{\boldsymbol{\Lambda}}_0\\
\dot{\boldsymbol{\lambda}}
\end{bmatrix}
=
\underbrace{\begin{bmatrix}
\boldsymbol{0}\\
\boldsymbol{G}_{\omega}\\
\boldsymbol{0}
\end{bmatrix}}_{\boldsymbol{B}_{\omega}}
\boldsymbol{\omega}_b
+
\underbrace{\begin{bmatrix}
\boldsymbol{0}\\
\boldsymbol{G}_{v}\\
\boldsymbol{0}
\end{bmatrix}}_{\boldsymbol{B}_{v}}
v_{\Vert}.
\end{equation}
Equation~\eqref{eq:block_diff_system} provides a linear mapping from the base inputs $(\boldsymbol{\omega}_b,v_{\Vert})$ to the time derivatives of the reduced implant state and the base insertion wrench, and will be used in the next subsection to compute the lateral-force sensitivity and derive the feedback law.

\subsection{Sensitivity computation}\label{sec:sensitivity}
Equation~\eqref{eq:block_diff_system} is a linear system that implicitly relates the base inputs $(\boldsymbol{\omega}_b,v_{\Vert})$ to the time derivatives $(\dot{\boldsymbol{q}},\dot{\boldsymbol{\Lambda}}_0,\dot{\boldsymbol{\lambda}})$.
By solving \eqref{eq:block_diff_system} at the current linearization point, we obtain an input--wrench sensitivity of the base insertion wrench:
\begin{equation}\label{eq:Lambda_dot_H}
\dot{\boldsymbol{\Lambda}}_0
=
\boldsymbol{H}_{\omega}\boldsymbol{\omega}_b
+
\boldsymbol{H}_{v}\,v_{\Vert},
\end{equation}
where $\boldsymbol{H}_{\omega}\in\mathbb{R}^{6\times 3}$ and $\boldsymbol{H}_{v}\in\mathbb{R}^{6\times 1}$ are the sensitivity matrices associated with the base angular velocity and the axial advancement speed, respectively.

Since $\boldsymbol{\Lambda}_0$ is expressed in the base body frame, its $y$--$z$ force components are directly the lateral components orthogonal to the insertion axis (the base $x$-axis).
We define
\begin{equation}\label{eq:fperp_def}
\boldsymbol{f}_{\perp}
=
\begin{bmatrix}
f_{0y}\\
f_{0z}
\end{bmatrix}
=
\boldsymbol{S}_{\perp}\boldsymbol{\Lambda}_0,
\qquad
\boldsymbol{S}_{\perp}=
\begin{bmatrix}
\boldsymbol{0}_{2\times4} & \mathbf{I}_2
\end{bmatrix}.
\end{equation}
Differentiating \eqref{eq:fperp_def} and using \eqref{eq:Lambda_dot_H} yields the lateral-force sensitivity:
\begin{equation}\label{eq:fperp_dot}
\dot{\boldsymbol{f}}_{\perp}
=
\boldsymbol{J}_{\perp}\boldsymbol{\omega}_b
+
\boldsymbol{b}_{\perp}\,v_{\Vert},
\end{equation}
with
\begin{equation}\label{eq:Jperp_bperp}
\boldsymbol{J}_{\perp}:=\boldsymbol{S}_{\perp}\boldsymbol{H}_{\omega}\in\mathbb{R}^{2\times 3},
\qquad
\boldsymbol{b}_{\perp}:=\boldsymbol{S}_{\perp}\boldsymbol{H}_{v}\in\mathbb{R}^{2\times 1}.
\end{equation}
This relationship will be used to design a continuous-time feedback law that suppresses the lateral force components during insertion.

\subsection{Lateral-force feedback law}\label{sec:feedback_law}
Our objective is to suppress the lateral components of the base insertion force, so that the propulsion is transmitted as much as possible along the implant axis and the risk of buckling is reduced.
Using the lateral-force sensitivity \eqref{eq:fperp_dot}, we impose a first-order stabilizing dynamics
\begin{equation}\label{eq:desired_dyn}
\dot{\boldsymbol{f}}_{\perp}=-k\,\boldsymbol{f}_{\perp},
\qquad k>0,
\end{equation}
where $k$ is a proportional gain. This choice enforces an exponential decay of $\boldsymbol{f}_{\perp}$ toward zero.

Substituting \eqref{eq:fperp_dot} into \eqref{eq:desired_dyn} gives the control equation
\begin{equation}\label{eq:control_equation}
\boldsymbol{J}_{\perp}\boldsymbol{\omega}_b
=
-k\,\boldsymbol{f}_{\perp}
-\boldsymbol{b}_{\perp}\,v_{\Vert}.
\end{equation}
Since $\boldsymbol{J}_{\perp}\in\mathbb{R}^{2\times 3}$ is non-square, we choose the minimum-norm solution using the pseudo-inverse:
\begin{equation}\label{eq:omega_dls}
\boldsymbol{\omega}_b
=
-\boldsymbol{J}_{\perp}^{\top}\Big(\boldsymbol{J}_{\perp}\boldsymbol{J}_{\perp}^{\top}
+\varepsilon\mathbf{I}_{2}\Big)^{-1}
\Big(k\,\boldsymbol{f}_{\perp}+\boldsymbol{b}_{\perp}\,v_{\Vert}\Big),
\quad \varepsilon>0,
\end{equation}
where $\varepsilon$ is the damping coefficient.

Under the local linearization assumption, and provided that $\boldsymbol{J}_{\perp}$ has full row rank, the proposed feedback yields the target closed-loop behavior \eqref{eq:desired_dyn}, driving the lateral force to zero. The commanded base angular velocity $\boldsymbol{\omega}_b$ continuously adjusts the insertion direction, whereas the axial speed $v{\Vert}$ guarantees forward progression while respecting the RCM-like constraint.

Overall, the insertion problem is formulated as lateral-force regulation under an RCM-like constraint at the cochlear entrance. Leveraging the differentiable, contact-aware implant model, we obtain an implicit sensitivity that maps $(\boldsymbol{\omega}_b, v{\Vert})$ to the time variation of the base insertion wrench and design a feedback law that suppresses lateral components without compromising axial advancement. In the following sections, we first validate the proposed model and then assess the resulting strategy through simulation and experimental insertion studies.

\section{Model validation}\label{section:experiment}
The primary aim of the proposed simulation is to predict implant deformation and interaction forces during insertion,
and to use this model to study how key parameters affect the implantation outcome. Accordingly, this section proceeds
in two steps. First, we validate the model by comparing simulations with benchtop insertion experiments performed at
different insertion angles. Second, we analyze the simulated force evolution to interpret the dominant mechanisms
governing the insertion process.

\begin{figure}[t]
	\centering
	\includegraphics[width=0.48\textwidth]{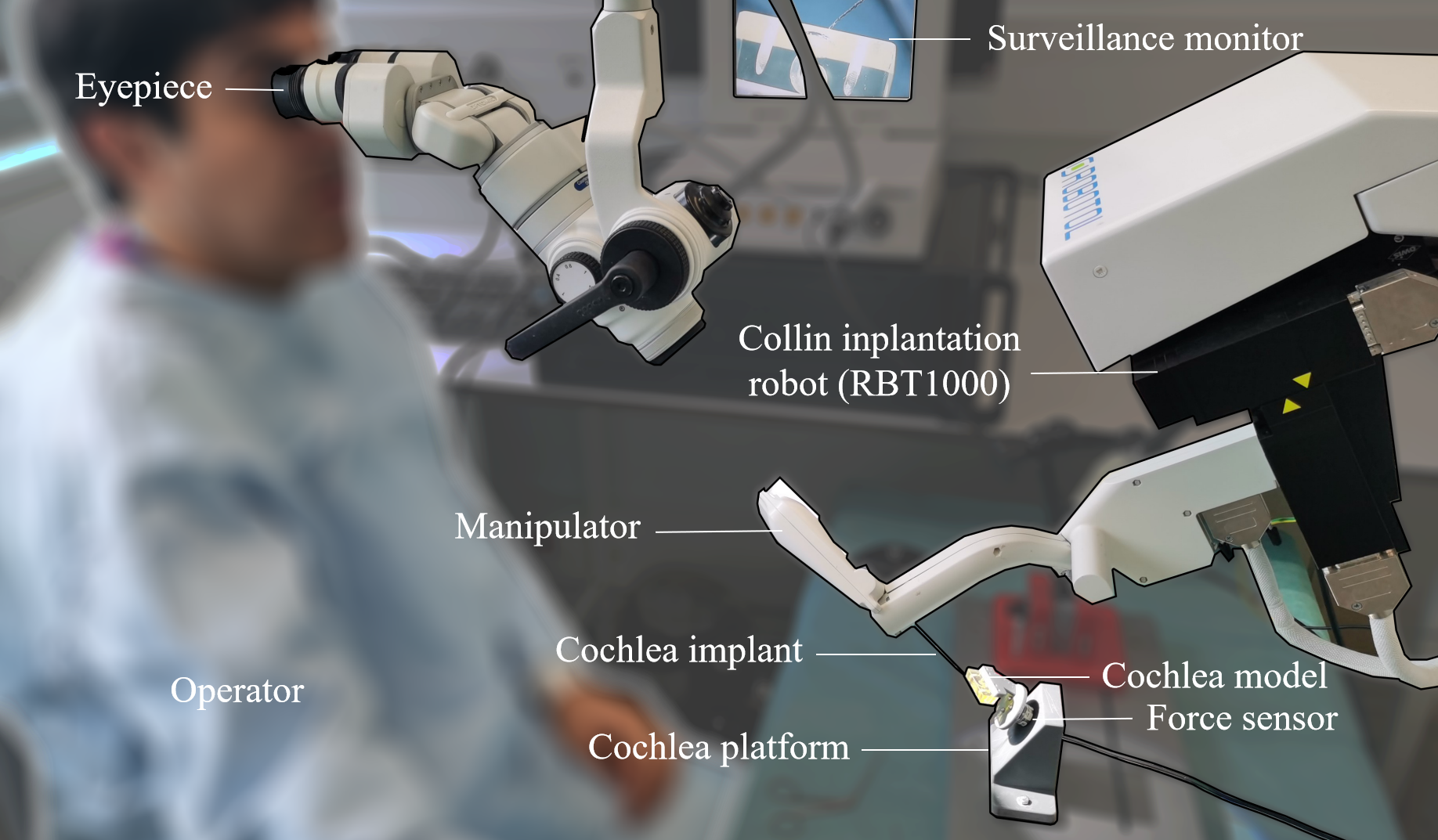}
	\caption{Robotic benchtop insertion platform. A Collin RobOtol (RBT1000) advances the implant into a
	cochlear phantom mounted on a 6-axis force/torque sensor.}
	\label{fig:collin}
\end{figure}
\begin{figure}[t]
	\centering
	\includegraphics[width=0.49\textwidth]{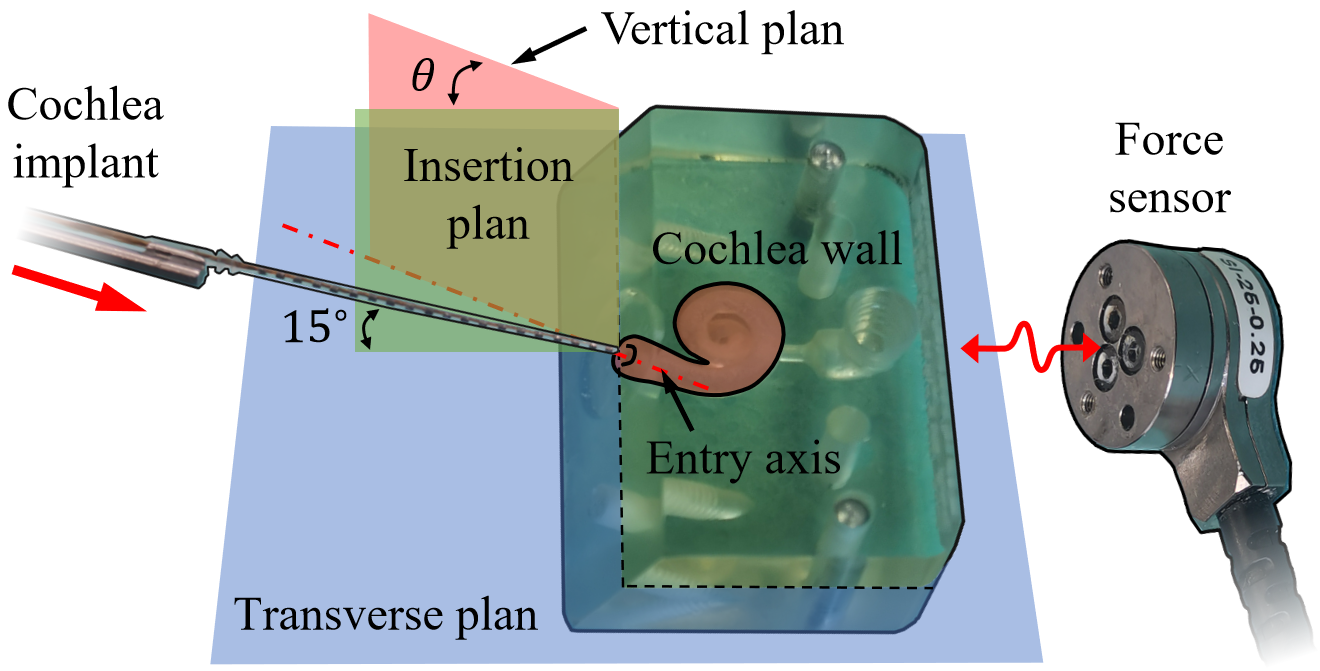}
	\caption{Definition of the insertion direction and angle parameters. The implant is advanced with a fixed
	pitch angle of $15^\circ$ and a variable yaw angle $\theta$ at the cochlear entrance. The cochlear phantom is
	mounted on a force/torque sensor to measure the transmitted interaction wrench.}
	\label{fig:setup1}
\end{figure}

\subsection{Experiment setup}
We conducted benchtop insertion experiments using a scaled resin 3D-printed cochlear phantom (OTICON Medical) as a
surrogate of the cochlea. Figures~\ref{fig:collin} and \ref{fig:setup1} respectively provide an overview of the
robotic platform and a detailed view of the mounting/sensing arrangement together with the definition of the
insertion angles.

The setup consists of the cochlear phantom and fixture, a 6-axis force/torque sensor (ATI Nano17), a robotic
insertion platform (Collin RobOtol, RBT1000), and the implant (Pasteur Institute). The measured implant
parameters are reported in Tab.~\ref{tab:Measured_parameters}. The friction coefficient $\mu$ is identified as $0.58$. During each trial, the implant is attached to the
robot probe. The phantom is rigidly mounted on the base, which is fixed to the force/torque sensor to measure the
interaction wrench transmitted through the cochlea fixture. The Nano17 sensor provides a force resolution of
0.318\,gf.

\begin{table}[t]
\caption{Experimental Parameters of the implant}
\label{tab:Measured_parameters}
\centering
\renewcommand{\arraystretch}{1.15}
\setlength{\tabcolsep}{6pt}
\begin{tabular}{l c l c}
\hline
\textbf{Parameter} & \textbf{Value} & \textbf{Parameter} & \textbf{Value} \\
\hline
Length $L$ & 25\,mm & Tip diameter & 0.3\,mm \\
Young's modulus $E$ & 25.2\,MPa & Base diameter & 0.4\,mm \\
\hline
\end{tabular}
\end{table}

\subsection{Insertion with different angles}
Prior clinical and in-vitro studies have shown that insertion outcome and force level depend on the insertion angle
at the cochlear entrance \cite{landsberger2015relationship,aebischer2021vitro}. In our setup, the pitch angle is
fixed to $15^\circ$ while the yaw angle $\theta$ is varied (Fig.~\ref{fig:setup1}). To evaluate the model under
different conditions, we consider three groups with $\theta\in\{0^\circ,10^\circ,20^\circ\}$ and perform paired
experiments and simulations for each case.

\begin{figure*}[p]
	\centering
	\subfigure[$0^\circ$ yaw]{\includegraphics[width=0.99\textwidth]{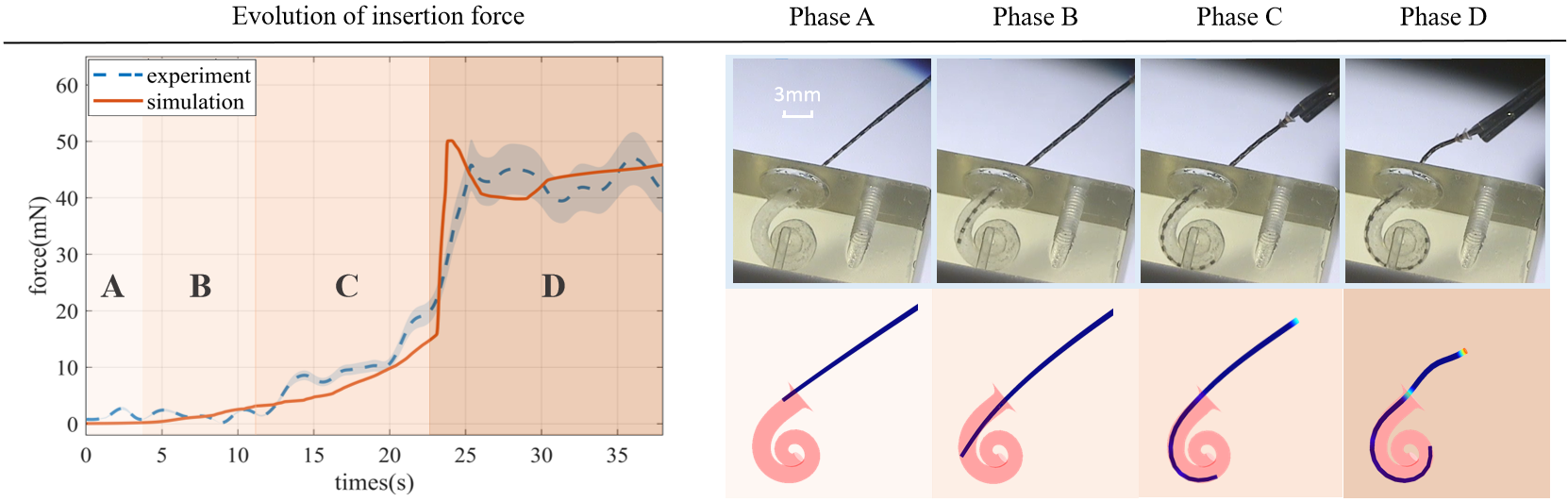}}
    \vspace{0pt}
	\subfigure[$10^\circ$ yaw]{\includegraphics[width=0.99\textwidth]{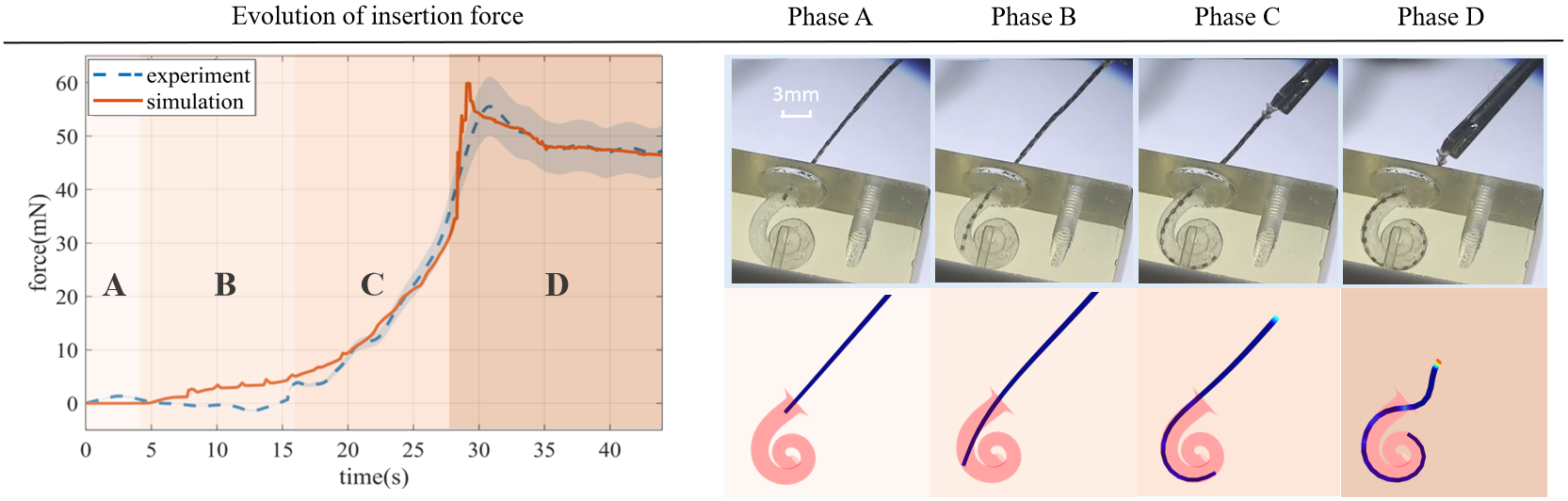}}
    \vspace{0pt}
	\subfigure[$20^\circ$ yaw]{\includegraphics[width=0.99\textwidth]{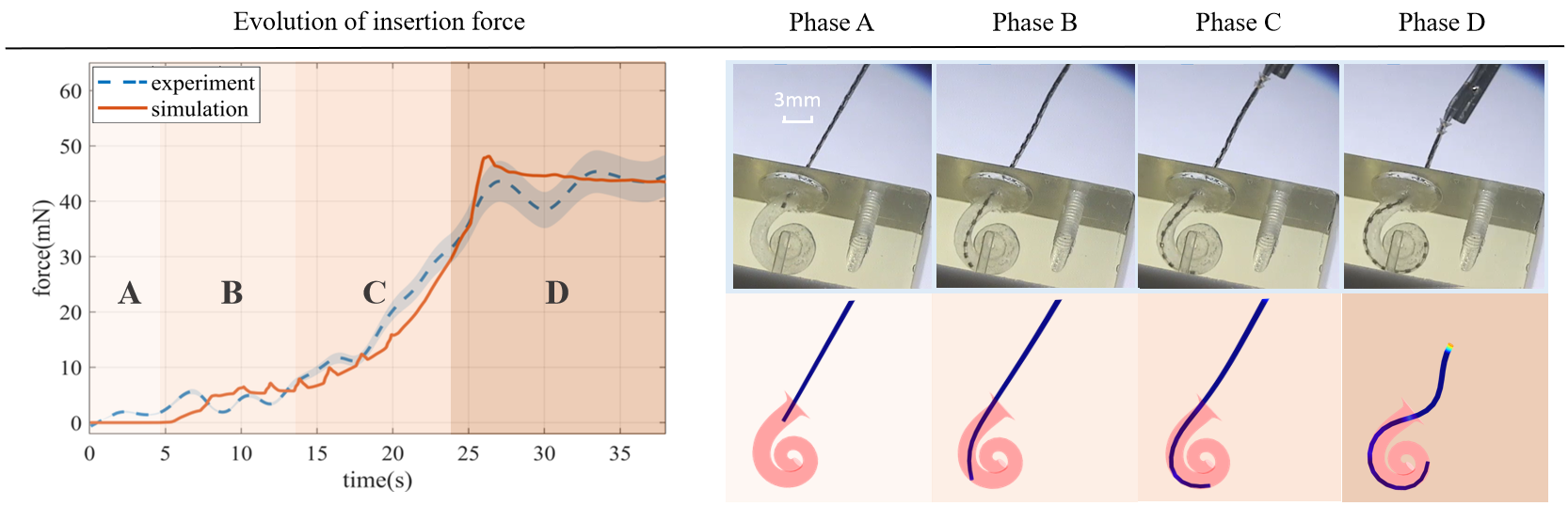}}
	\caption{Experiment--simulation comparison for three insertion yaw angles $\theta\in\{0^\circ,10^\circ,20^\circ\}$
	(with fixed pitch $15^\circ$).}
	\label{fig:8}
	\vspace{0.01in}
\end{figure*}

To isolate the effect of $\theta$, the axial advancement speed is fixed to 1\,mm/s in all trials. Each insertion is
terminated when forward progression is no longer possible (i.e., the implant becomes locked or buckles), thereby
defining the achieved insertion depth. Representative experiment/simulation comparisons are shown in Fig.~\ref{fig:8}.

\subsection{Experimental result}\label{sec:expec}
Figure~\ref{fig:8} compares representative insertion sequences from experiments and simulations for
$\theta=0^\circ$, $10^\circ$, and $20^\circ$.

\subsubsection{Implantation process analysis}
The insertion process can be divided into four stages based on the observed contact pattern.
In phase~A, the implant advances with negligible interaction.
In phase~B, the tip first contacts the inner wall and slides along it, leading to localized deformation and discrete
point contacts.
In phase~C, the implant progressively conforms to the lumen curvature and a finite contact region develops, resulting
in distributed contact and increasing friction. As the implant is advanced further, the decreasing lumen radius and
the growing contact area jointly increase elastic resistance and friction, leading to a rapid nonlinear rise of the
required insertion force.
Finally, in phase~D, the implant can no longer sustain the pushing load: a suspended segment buckles while parts of
the contact region become self-locking due to the change of contact orientation, and forward progression stops.

\subsubsection{Performance of the model}
We evaluate the model accuracy in terms of (i) global deformation shape, (ii) contact pattern (localized vs.
distributed), (iii) insertion-force evolution, and (iv) onset of buckling/locking. Overall, the simulation closely
matches experimental observations, including the transition from point contact (phase~B) to distributed contact
(phase~C) and the subsequent buckling behavior (phase~D).

To quantify the agreement in force magnitude, we introduce the integrated mean error (IME),
\[
\mathrm{IME}=\int_{0}^{T}\left|\frac{F_{\mathrm{exp}}(t)-F_{\mathrm{sim}}(t)}{F_{\mathrm{exp}}(t)}\right|\mathrm{d}t,
\]
where $F_{\mathrm{exp}}$ and $F_{\mathrm{sim}}$ denote the measured and simulated insertion-force magnitudes,
respectively. As shown in Fig.~\ref{fig:72}, the simulated force reproduces the experimental trend with
$\mathrm{IME}=16.3\%$.

\subsubsection{Influence of the insertion angle}
Both experiments and simulations indicate that the insertion yaw angle $\theta$ primarily affects when the system
enters the locking/buckling regime (phase~D), and thus the achievable insertion depth. In our setup,
$\theta=10^\circ$ yields the deepest insertion among the three tested conditions. More generally, the optimal angle
depends on the cochlear lumen geometry and the implant mechanical properties, which motivates the use of simulation to
evaluate candidate angles before insertion and reduce the risk of premature locking or buckling.

\section{Path Planning Validation}\label{section:experiment2}
\subsection{Protocol and metrics}\label{sec:pathplan_protocol}
We validate the proposed path-planning method in simulation by comparing two insertion strategies: (i) a constant-path strategy, where the base orientation is kept fixed throughout the insertion, and (ii) an optimized-path strategy, where the base orientation is updated according to the planned sequence as a function of insertion depth.

Insertion depth is reported as the cochlear spiral rotation angle~$\alpha$ (in degrees). The implant tip is mapped to the cochlear spiral, and the corresponding spiral angle~$\alpha$ is used as the insertion-depth measure. Insertion forces are reported as the resultant of the measured 3-axis forces.
Unless otherwise stated, we report the maximum insertion depth~$\alpha_{\max}$ together with representative trajectory and contact behaviors to characterize feasibility during deep insertion.

\subsection{Initial state and parameter settings}\label{sec:pathplan_init}
For the numerical study of path planning, the material and geometric parameters of the implant, as well as the geometric parameters of the cochlea, are kept consistent with the parameters described in the model validation Section~\ref{section:experiment}. Starting from these calibrated parameters, the optimization algorithm iterates to compute a depth-indexed base-motion sequence for insertion.

\begin{figure}[h]
	\centering
	\includegraphics[width=0.39\textwidth]{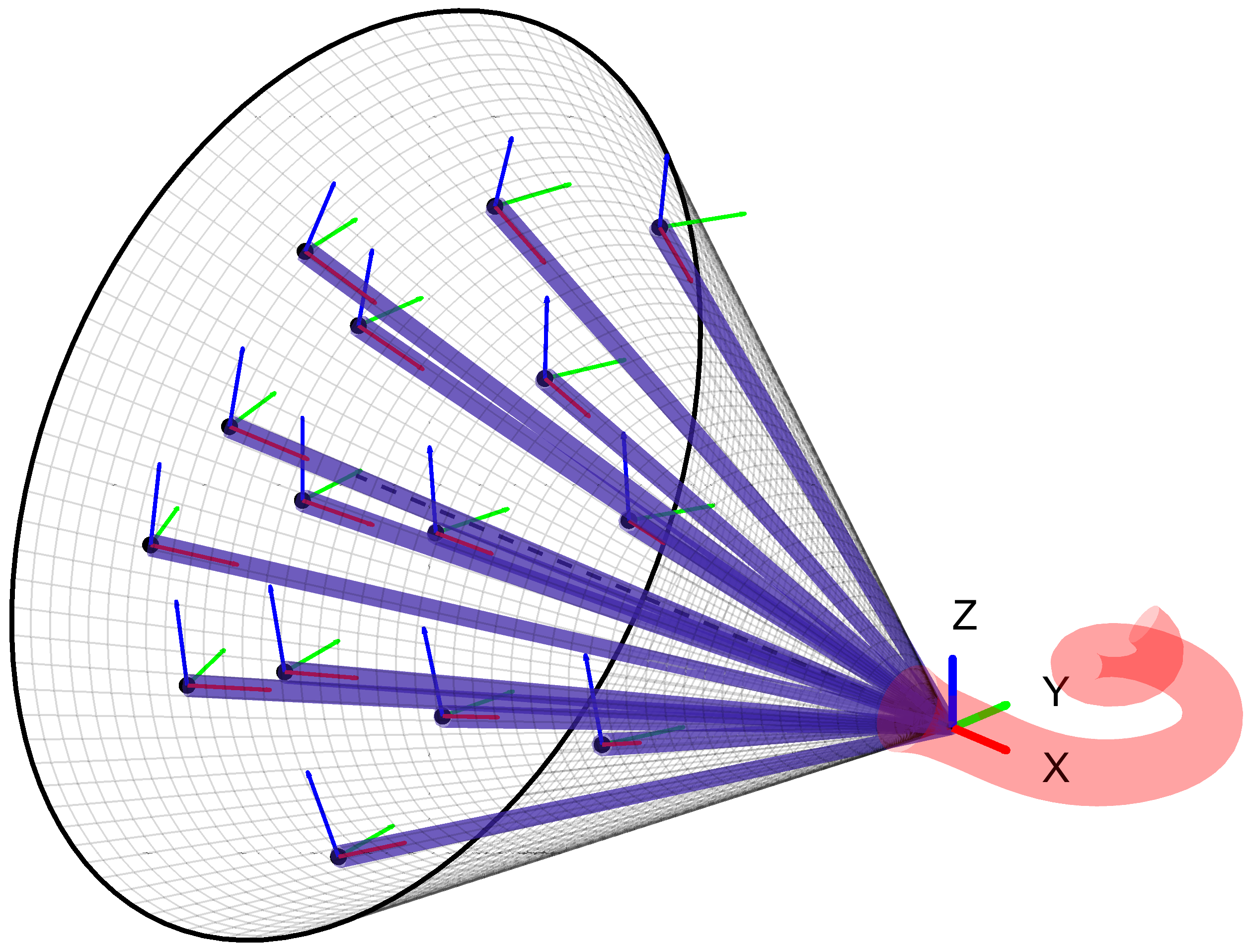}
	\caption{Initialization of insertion orientations. Reference orientations are sampled on a spherical cone whose axis is aligned with the cochlear-entry centerline; each sample defines an initial base frame (triad) for the planner.}
	\label{fig:init_sampling}
\end{figure}

The forward advancement step is set to $0.05~\mathrm{mm}$. To assess robustness with respect to the initial alignment, we initialize the planner from multiple reference orientations sampled on a spherical cone whose axis is aligned with the cochlear-entry centerline, as illustrated in Fig.~\ref{fig:init_sampling}. This sampling provides a broad set of initial insertion directions, ranging from near-axis configurations to large angular deviations. 
\begin{figure*}[t]
	\centering
	\includegraphics[width=0.8\textwidth]{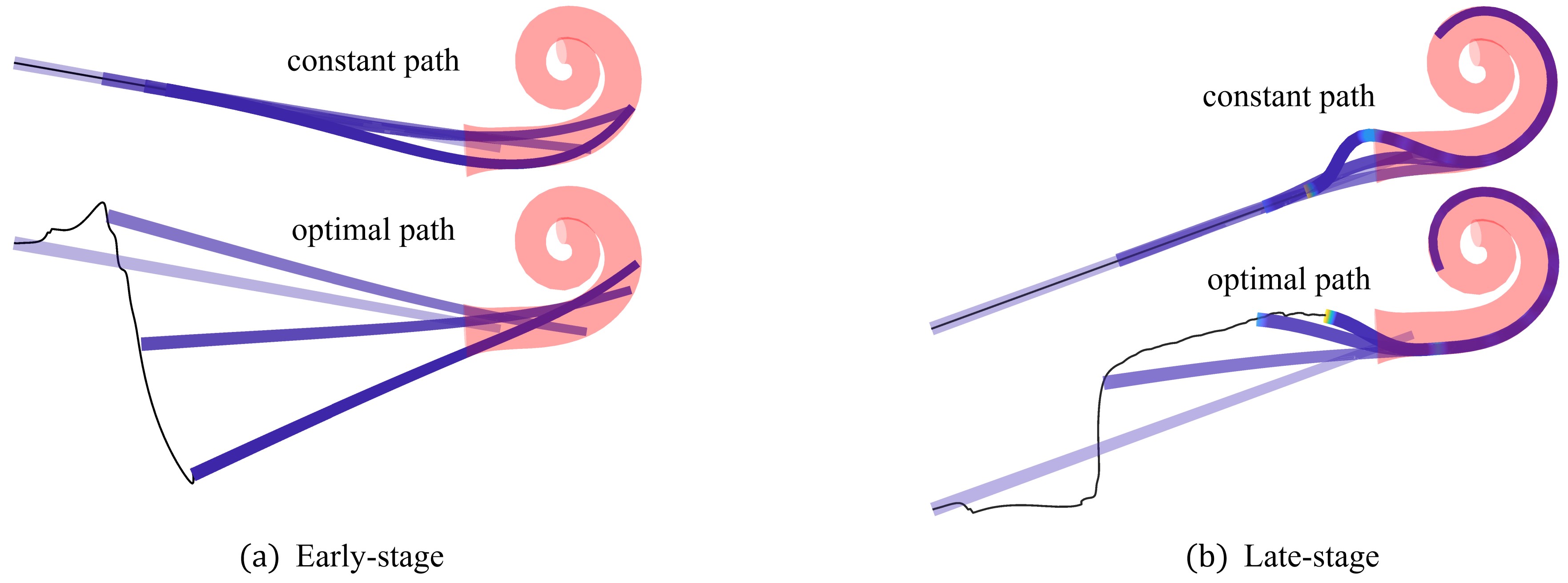}
	\caption{Simulation comparison between the constant-path and optimized-path insertion strategies under two representative initial orientations.
	(a) Early-stage behavior: the optimized-path strategy reduces excessive contact with the cochlear lumen compared with the constant-path strategy.
	(b) Late-stage behavior: the constant-path strategy exhibits an early buckling/instability, whereas the optimized-path strategy delays and mitigates buckling, enabling deeper advancement.
	The black curve depicts the insertion path of the implant base.}
	\label{fig:compare_buckling}
\end{figure*}
\subsection{Simulation results: optimal path characteristics}\label{sec:pathplan_sim}
In simulation, the optimized insertion trajectory typically exhibits three characteristic stages, reflecting the progressive activation of contact constraints:
\begin{enumerate}
	\item {Free-motion stage.}
	At the beginning of insertion, the implant does not contact the cochlear wall. As a result, no active contact constraints are present and the sensitivity of the objective/constraints with respect to the base orientation is negligible. During this phase, the base orientation remains approximately constant while the implant advances forward.
	\item {Contact-driven adjustment stage.}
	As the implant starts contacting the cochlear inner wall, contact/constraint reactions emerge and the optimization updates the base orientation accordingly, following the descent direction prescribed by the proposed planning rule. In this stage, the base orientation varies with insertion depth to limit unnecessary wall interaction and maintain feasible advancement under growing contact constraints.
	\item {Limit/instability stage.}
	At deep insertion, a limit is reached when further orientation adjustment can no longer make the contact reactions compatible with the forward advancement direction. The insertion then becomes compression-dominated and prone to instability, which manifests as buckling and ultimately prevents further advancement.
\end{enumerate}

The complementary roles of early contact mitigation and late-stage stability are illustrated in Fig.~\ref{fig:compare_buckling}. 
In Fig.~\ref{fig:compare_buckling}(a), the optimized-path strategy steers the implant toward a more compatible direction during the early insertion stage, thereby reducing excessive contact with the cochlear lumen compared with the constant-path insertion. 
In Fig.~\ref{fig:compare_buckling}(b), once contact dominates the mechanics at deep insertion, the constant-path strategy cannot maintain feasible advancement and buckles prematurely, whereas the optimized-path strategy adapts the base orientation to remain compatible with the contact constraints, thereby delaying buckling and allowing further advancement.

\begin{figure}[t]
	\centering
	\includegraphics[width=0.4\textwidth]{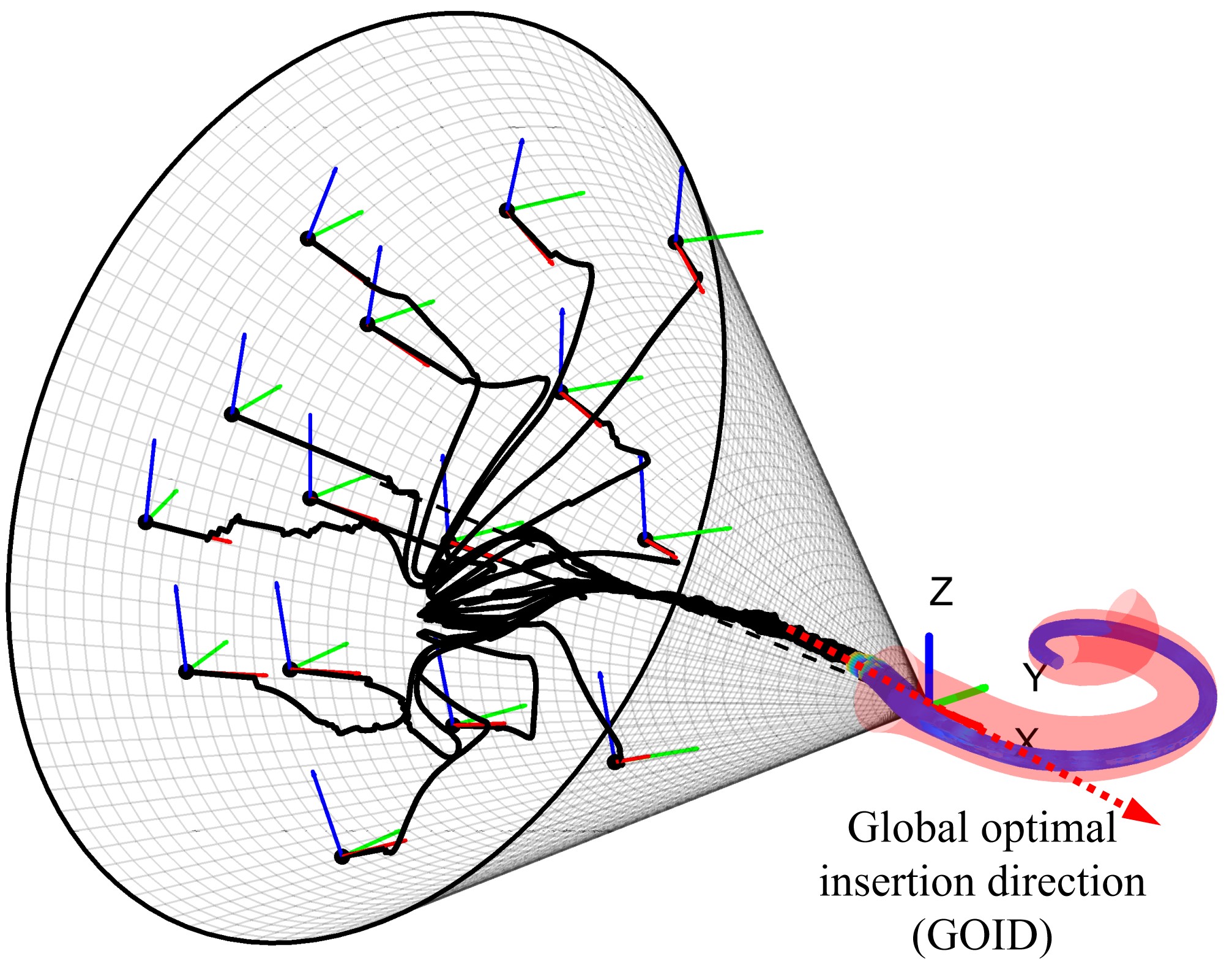}
	\caption{Optimized insertion trajectories from multiple initial orientations sampled on a cone around the cochlear-entry axis. 
	All trajectories rapidly adjust in the early stage and converge in the late stage; the resulting global optimal insertion direction (GOID) is shown by the red arrow.}
	\label{fig:GOID}
\end{figure}

Beyond representative cases, Fig.~\ref{fig:GOID} summarizes the optimized insertion trajectories obtained from the full set of initial orientations sampled on the spherical cone. 
Even for initial directions far from the cochlear-entry axis, the optimized trajectories rapidly adjust in the early stage toward a feasible orientation. 
As insertion proceeds, the trajectories progressively converge and become nearly overlapped at deep insertion, indicating that the proposed planning procedure consistently funnels a broad range of initial configurations toward a common strategy.

Based on this convergence property, we define the global optimal insertion direction (GOID) as the mean direction of the late-stage segments of all converged optimized trajectories, which is depicted by the red arrow in Fig.~\ref{fig:GOID}. 
This well-defined GOID provides a compact, global characterization of the preferred insertion direction implied by the planner and will be used in the following section to guide the experimental validation.

Given the agreement observed in Section~\ref{section:experiment}, we replay the optimized base-motion sequence computed in simulation in an open-loop manner to assess sim-to-real transferability under a controlled setup. In practice, the implant shape is difficult to measure reliably during insertion; therefore, the experiments are performed in open loop by executing pre-computed motion sequences. The following experiments demonstrate consistent improvements across repeated trials and different initial orientations.

\begin{figure}[t]
	\centering
	\includegraphics[width=0.45\textwidth]{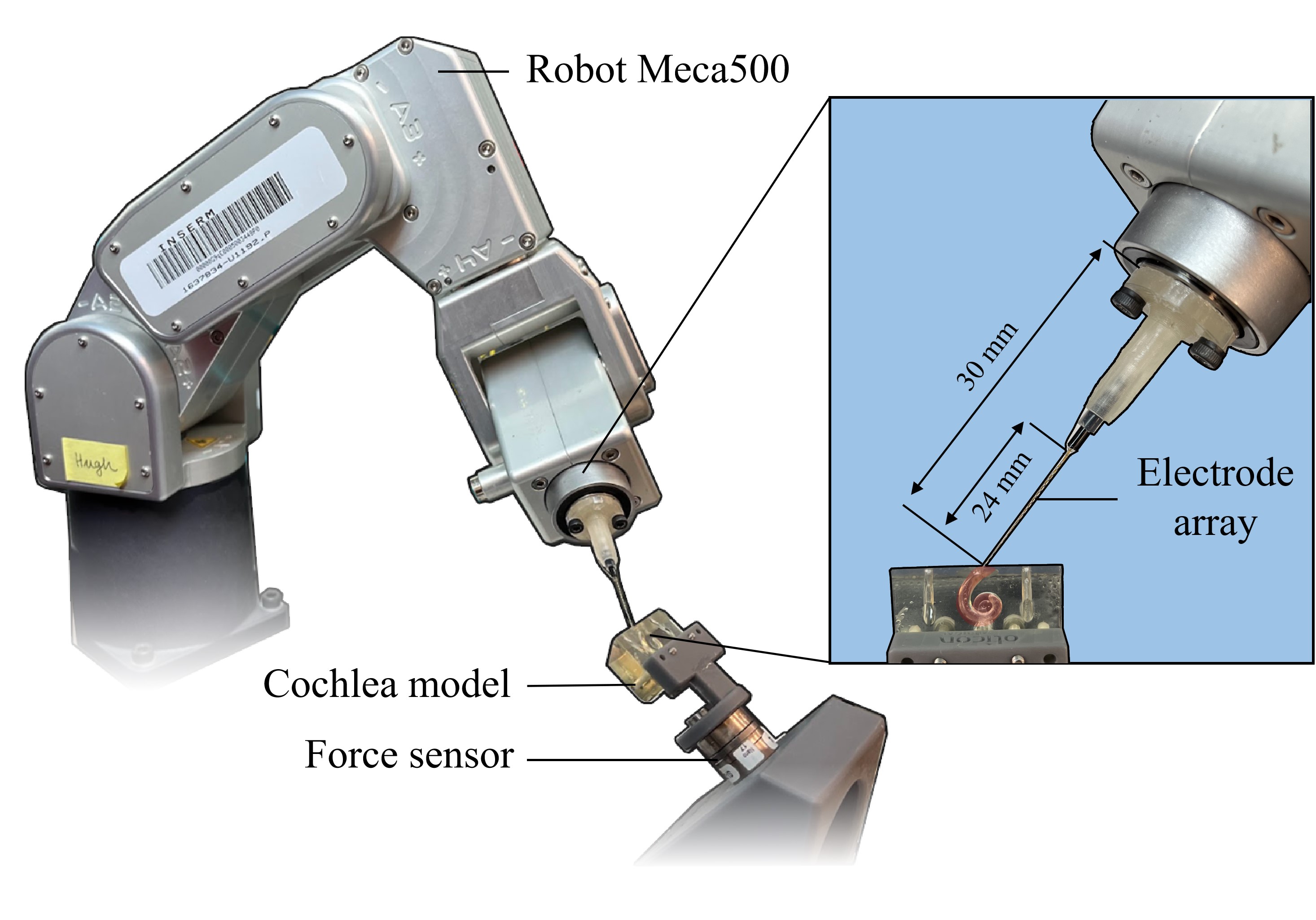}
	\caption{Experimental setup for cochlear-implant electrode insertion. A Meca500 robot drives the electrode array into a cochlea model while the interaction force is measured by an inline force sensor.}

	\label{fig:meca500}
\end{figure}

\subsection{Experiment setup}\label{sec:pathplan_setup}
In the experimental phase of this study, we employed a scaled resin 3D cochlea model (provided by OTICON Medical) as a substitute for the actual cochlea. The experimental setup, shown in Fig.~\ref{fig:meca500}, comprised a cochlea fixture, a six-degree-of-freedom force sensor, a Meca500 robot (6-DoF end-effector), and the implant.

During insertion, the implant is fixed on the probe attached to the robot end-effector, while the cochlea model is secured on the base. A six-degree-of-freedom force sensor (NANO17, ATI Industrial Automation; resolution down to 0.318 gram-force) is installed beneath the cochlea placement to measure insertion forces.

\subsection{Experimental results}\label{sec:pathplan_exp}
\begin{figure}[t]
	\centering
	\includegraphics[width=0.48\textwidth]{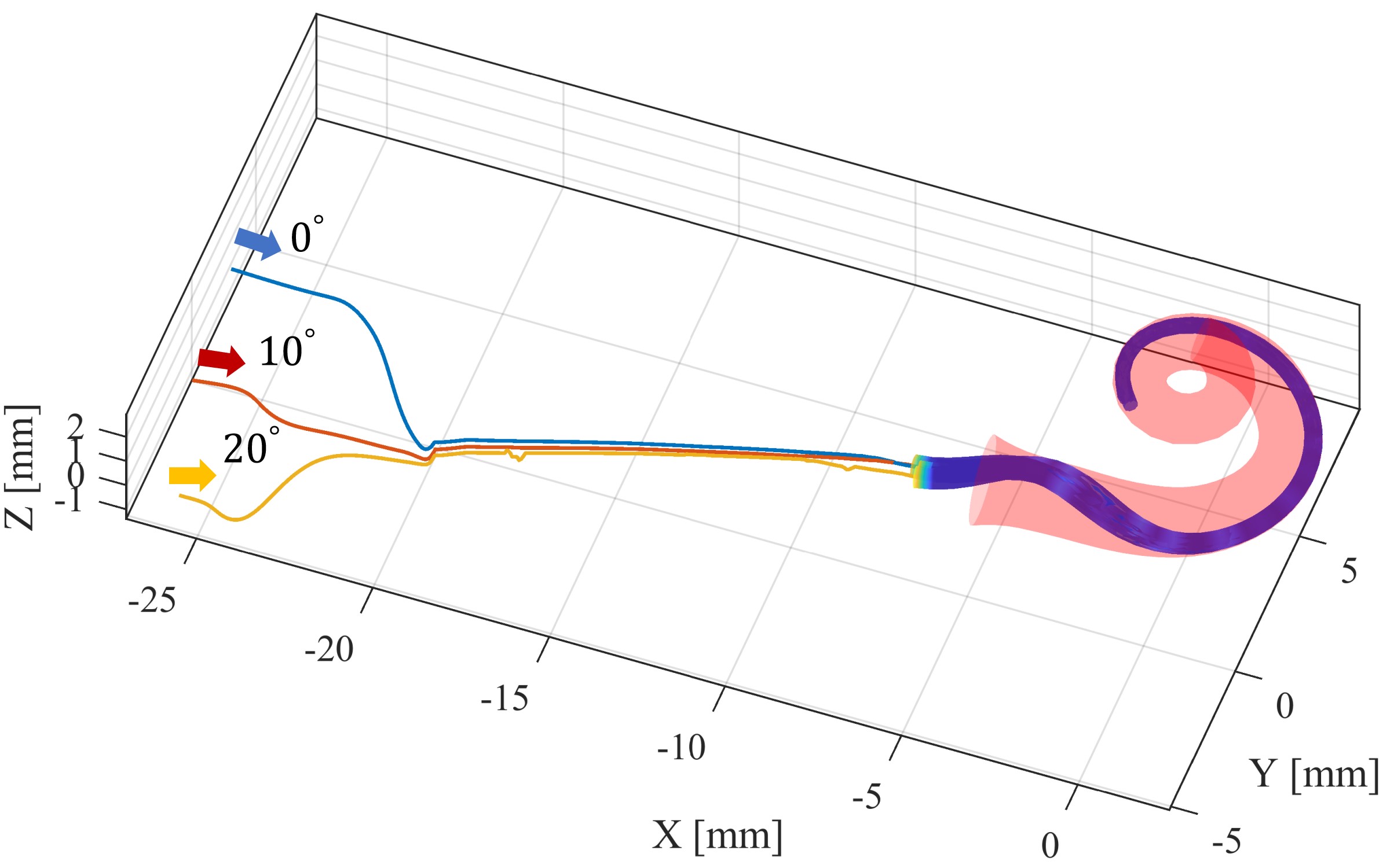}
	\caption{Optimized insertion paths used for the experiments, computed in simulation from three initial orientations defined with respect to the GOID: $0^\circ$ (GOID), $10^\circ$, and $20^\circ$.}
	\label{fig:path}
\end{figure}

To reflect practical implantation conditions, we evaluate three initial orientations in experiments. Specifically, We take the global optimal insertion direction (GOID) identified in the previous simulation study as the reference and denote it as the $0^\circ$ case. 
The other two initial orientations are generated by rotating the GOID such that, in the global frame, the resulting insertion direction forms an angle of $10^\circ$ and $20^\circ$ with the global $Z$-axis, respectively.  For each initial orientation, we compute the corresponding optimized insertion path in simulation and directly replay the resulting base-motion sequence on the robot. The three optimized paths used for the experiments are shown in Fig.~\ref{fig:path}.

\begin{figure}[t]
	\centering
	\includegraphics[width=0.49\textwidth]{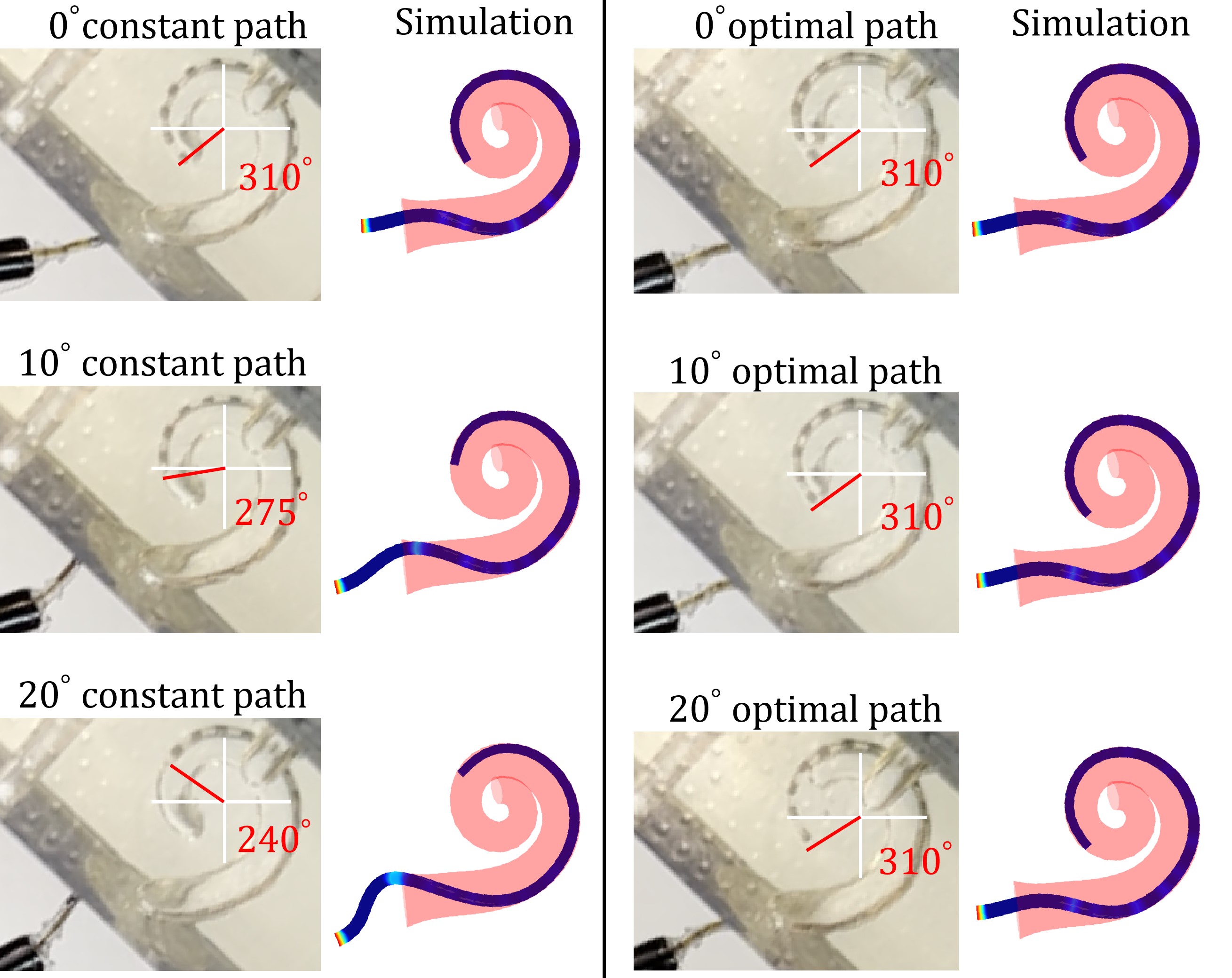}
	\caption{Maximum insertion depth for constant-path insertion and optimized-path insertion under different initial orientations ($0^\circ$, $10^\circ$, $20^\circ$), defined with respect to the GOID.}
	\label{fig:pathex}
\end{figure}

After executing these open-loop motion sequences, Fig.~\ref{fig:pathex} reports the maximum insertion depth~$\alpha_{\max}$ achieved under the constant-path strategy and the optimized-path strategy. Each condition is repeated five times under identical settings.

Under the constant-path strategy, the maximum insertion depth depends on the initial orientation: the deepest insertion is achieved near the GOID ($0^\circ$), and the maximum depth decreases as the angular offset increases. In contrast, the optimized-path strategy consistently reaches approximately $310^\circ$ across all three initial orientations, indicating improved robustness with respect to initial misalignment.

\begin{figure}[h]
	\centering
	\subfigure[Total insertion force]{%
		\includegraphics[width=0.495\textwidth]{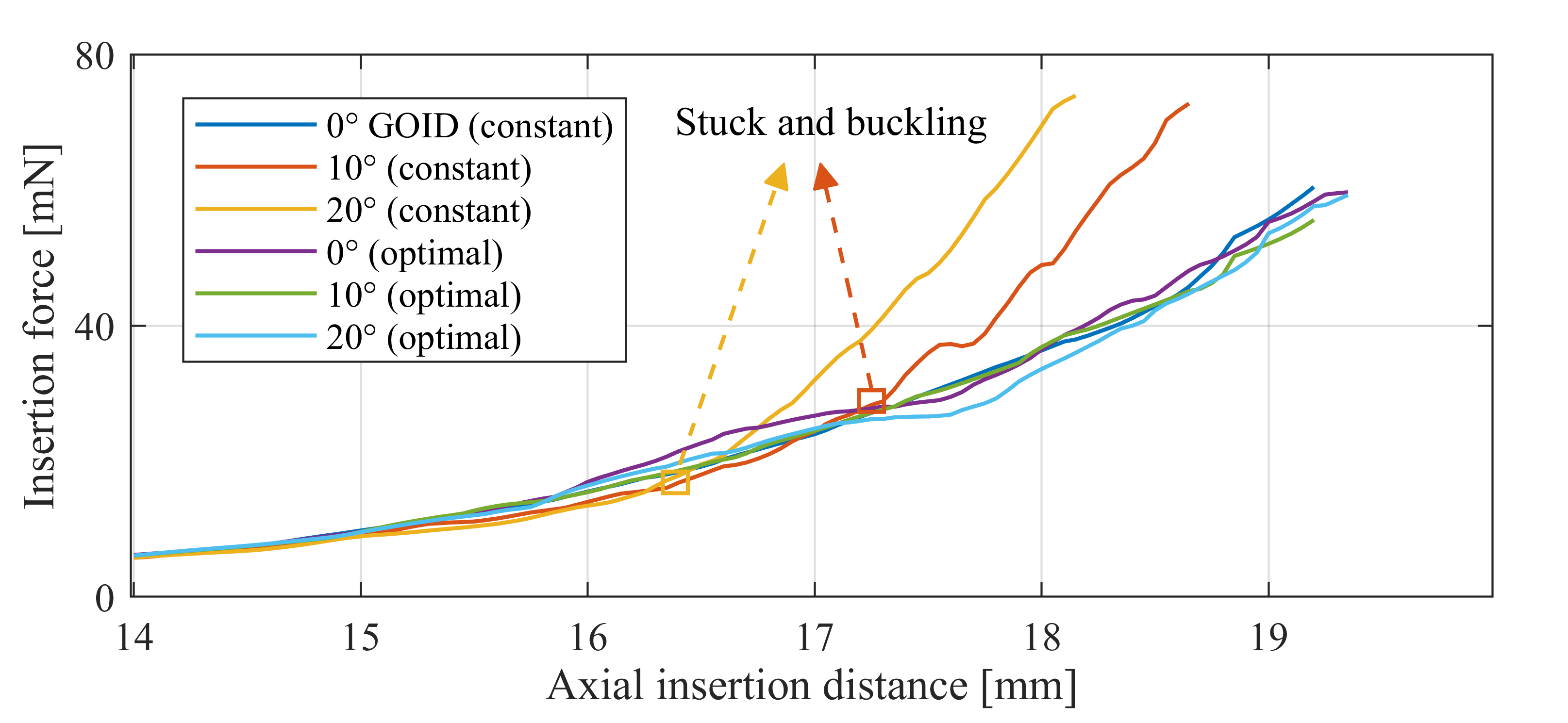}
		\label{fig:72a}
	}
	\hfill
	\subfigure[Lateral insertion force]{%
		\includegraphics[width=0.495\textwidth]{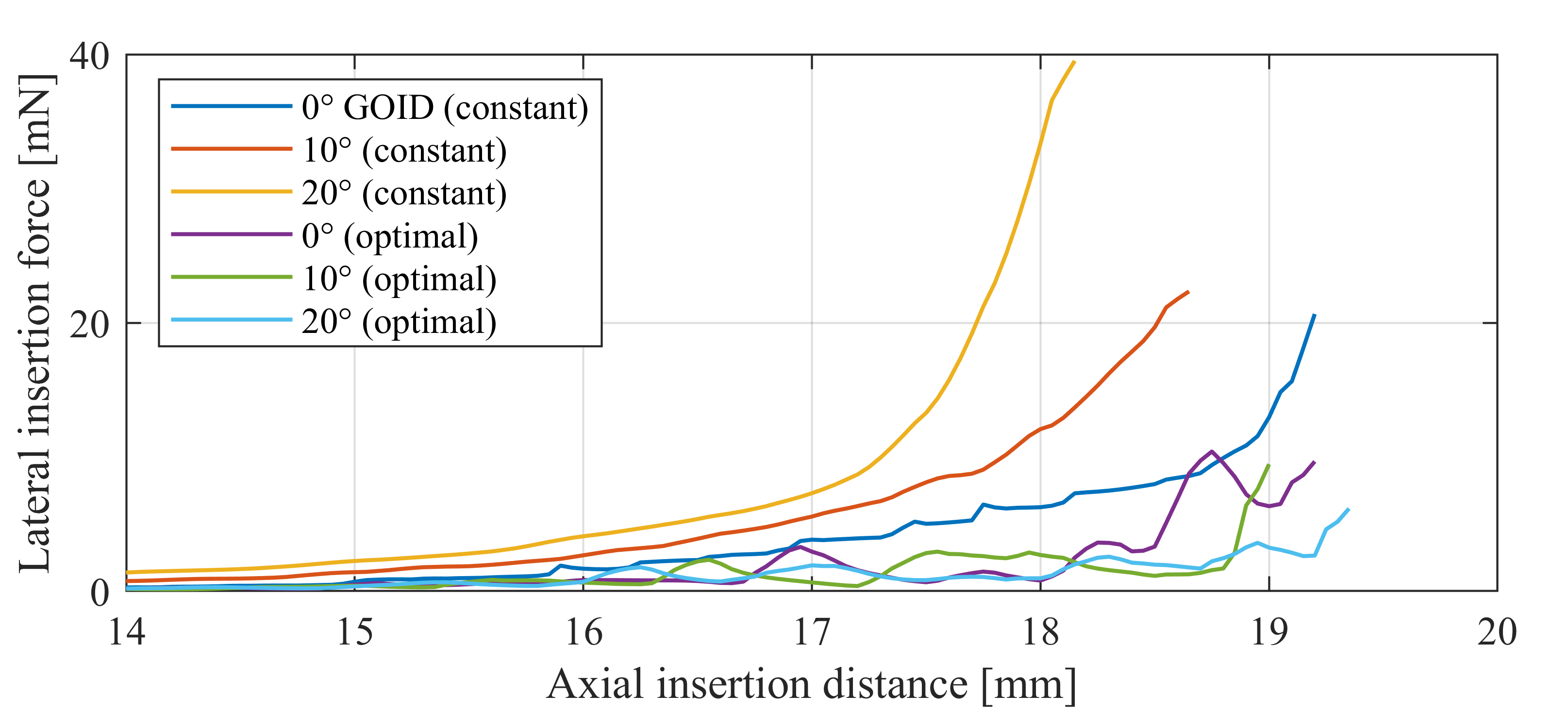}
		\label{fig:72b}
	}
	\caption{Insertion-force measurements for different cases starting from different initial orientations (defined with respect to the GOID).}
	\label{fig:72}
\end{figure}

Moreover, the insertion-force measurements provide an additional validation signal beyond the final insertion depth.
As shown in Fig.~\ref{fig:72}(a), the constant-path insertions with initial orientations of $10^\circ$ and $20^\circ$
exhibit an early stuck/buckling event, occurring at axial insertion distances of approximately $16.3~\mathrm{mm}$ and
$17.3~\mathrm{mm}$, respectively. Around these depths, the force curve shows a clear regime change: the force rising
slope increases abruptly, indicating a sudden loss of mechanical feasibility associated with excessive contact and
compression-dominated loading. In contrast, for the GOID case ($0^\circ$), both strategies (constant-path and
optimized-path) avoid premature stuck/buckling.

\begin{figure*}[t]
	\centering
	\includegraphics[width=0.95\textwidth]{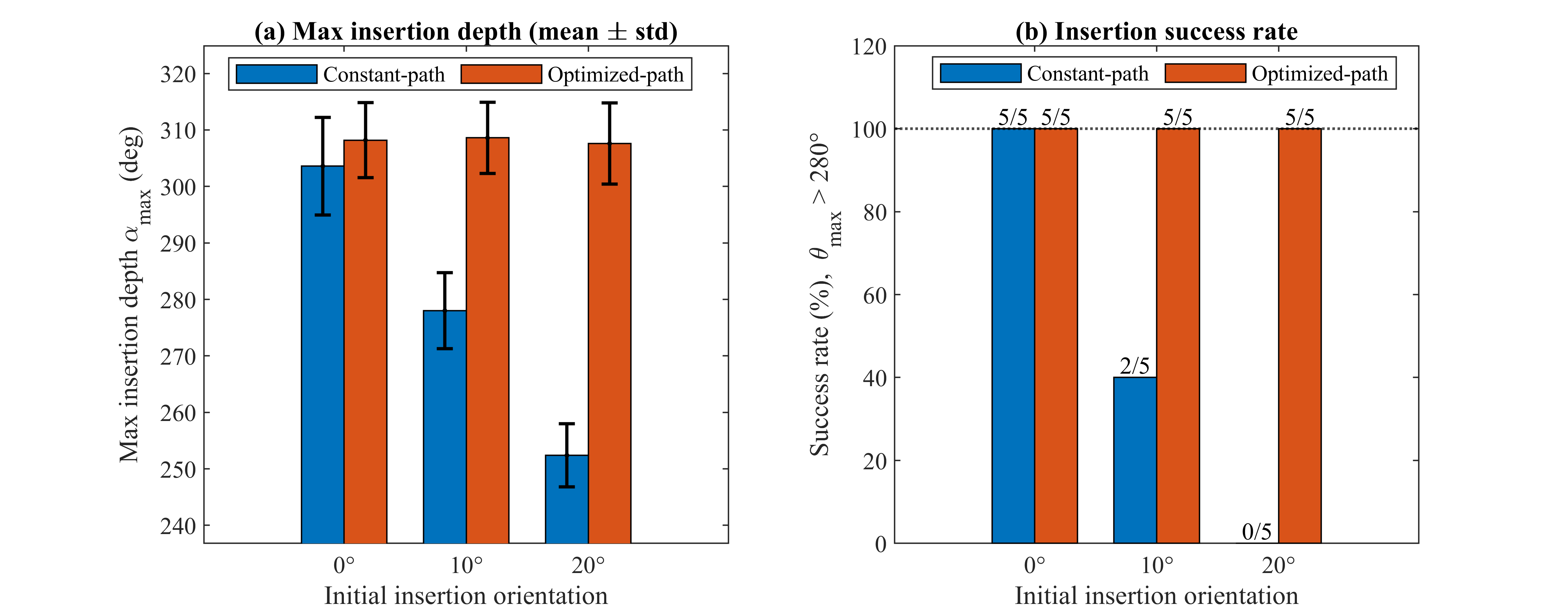}
	\caption{Statistical comparison over repeated trials (n=5). Left: mean maximum insertion depth (mean $\pm$ std). Right: insertion success rate (successful if $\alpha_{\max}\ge 280^\circ$).}
	\label{fig:stats}
\end{figure*}

Beyond the resultant force magnitude, Fig.~\ref{fig:72}(b) reports the lateral insertion force, which directly
reflects off-axis interaction that may induce bending and buckling. The optimized-path strategy effectively
suppresses the lateral force and yields consistently lower lateral-force levels than the constant-path strategy
across all initial orientations. Consequently, the optimized-path strategy produces mutually consistent force
profiles across different initial orientations, with smoother force evolution and reduced sensitivity to initial
misalignment, which complements the depth statistics and further supports the robustness of the proposed planning
method.

To quantify repeatability, we summarize the repeated trials using (i) the mean and standard deviation of the maximum insertion depth~$\alpha_{\max}$ and (ii) the insertion success rate. A trial is counted as successful if $\alpha_{\max} \ge 280^\circ$ (achieve insertion of at least 18 electrodes.) without buckling-induced premature termination. Fig.~\ref{fig:stats} and Tab.~\ref{tab:repeatability} provides a quantitative comparison over repeated trials (n=5) under three initial orientations. 
For the constant-path strategy, the maximum insertion depth decreases markedly as the initial orientation deviates from the GOID, and the trial-to-trial variability increases; this degradation is also reflected by a sharp drop in the success rate (from 5/5 at $0^\circ$ to 2/5 at $10^\circ$ and 0/5 at $20^\circ$). 
In contrast, the optimized-path strategy consistently achieves deep insertion close to $310^\circ$ with smaller dispersion across all three initial orientations, and maintains a 100\% success rate (5/5) for every case. 
Overall, these statistics confirm that the proposed planning method not only improves the average achievable insertion depth, but also substantially enhances robustness and repeatability with respect to initial misalignment.

\begin{table}[t]
  \centering
  \caption{Repeatability metrics over $n=5$ trials under different initial conditions. A trial is successful if $\alpha_{\max}\ge 280^\circ$ without premature termination.}
  \label{tab:repeatability}
  \setlength{\tabcolsep}{6pt}
  \begin{tabular}{@{}ll
                  S[table-format=3.1]  
                  S[table-format=2.1]  
                  S[table-format=1.0]  
                  S[table-format=1.0]  
                  @{}}
    \toprule
    \textbf{Strategy} & \textbf{Condition} &
    \multicolumn{2}{c}{\textbf{$\alpha_{\max}$ (deg)}} &
    \multicolumn{1}{c}{\textbf{Succ.}} &
    \multicolumn{1}{c}{\textbf{Premature}} \\
    \cmidrule(lr){3-4}
    \cmidrule(lr){5-6}
    & & \multicolumn{1}{c}{Mean} & \multicolumn{1}{c}{Std} &
    \multicolumn{1}{c}{$N_{\mathrm{succ}}/5$} &
    \multicolumn{1}{c}{$N$} \\
    \midrule
    Constant-path  & $0^\circ$   & 305.0 &  17 & 5 & 0 \\
                   & $10^\circ$  & 278.0 & 15.0 & 2 & 3 \\
                   & $20^\circ$     & 253.0 & 10.0 & 0 & 5 \\
    \addlinespace
    Optimized-path & $0^\circ$   & 309.5 &  12 & 5 & 0 \\
                   & $10^\circ$  & 308.0 &  10 & 5 & 0 \\
                   & $20^\circ$     & 307.5 &  13 & 5 & 0 \\
    \bottomrule
  \end{tabular}
\end{table}

In summary, the experiments corroborate the simulation findings: the proposed path-planning method increases the achievable insertion depth and improves robustness with respect to initial insertion orientation, while being directly executable in open loop. Incorporating closed-loop force feedback is a natural extension to further improve reliability under broader variability.

\section{Conclusion}\label{section:conclusion}
This paper presented a unified image-to-simulation pipeline for cochlear-implant insertion, integrating CT-derived patient-specific anatomy, contact-aware mechanical modeling, and robotic path planning. We developed a low-dimensional Cosserat-rod model of the electrode array that captures bending, torsion, shear, and distributed frictional contact, regularized to ensure well-posed and temporally continuous evolution across stick–slip transitions. To enable efficient and differentiable contact queries, we introduced a patient-specific analytic parameterization of the scala-tympani lumen derived from CT imaging, with closest-point contact-pair construction performed directly on the continuous surface.

Building on a differentiated equilibrium-constraint framework, insertion path optimization was formulated as a continuous-time, contact-aware direction adaptation problem under an RCM-like constraint at the cochlear entrance. This produces an online, feedback-based direction-update law that regulates lateral insertion forces while maintaining prescribed axial advancement, supporting predictive, safety-aware robotic control. Simulation and benchtop experiments with varying insertion angles validated the framework, reproducing electrode deformation and force evolution and revealing the mechanisms underlying locking and buckling.

Aligned with the goals of robot-assisted medical imaging, this work demonstrates how CT-based anatomy can enhance both modeling fidelity and robotic planning, enabling precise, autonomous, and safety-conscious interventions. Future work will focus on closed-loop robotic insertion by integrating intraoperative imaging and force sensing with the contact-aware update law. Additionally, we plan to extend the framework to account for external anatomical constraints, supporting path planning and autonomous control under broader surgical workspace limitations.

\section{Appendix}
\subsection{Lie group notations}\label{notations}
The adjoint representation of the Lie algebra is given by	
$$
{\rm{ad}}_{\boldsymbol{\xi}}= \left(\begin{matrix}
	\tilde{\boldsymbol{\kappa}}&\boldsymbol{0}_{3\times3}\\\tilde{\boldsymbol{\epsilon}}&\tilde{\boldsymbol{\kappa}}
\end{matrix}\right)\in \mathbb{R}^{6\times6}, \ {\rm{ad}}_{\boldsymbol{\eta}}= \left(\begin{matrix}
	\tilde{\boldsymbol{w}}&\boldsymbol{0}_{3\times3}\\\tilde{\boldsymbol{v}}&\tilde{\boldsymbol{w}}
\end{matrix}\right)\in \mathbb{R}^{6\times6},$$
where the operator $\tilde{(\cdot)}$ represents a conversion from a 3-dimensional vector to its skew-symmetric matrix.
\subsection{Transformation matrix}\label{notations2}
The matrix transforming the velocity or acceleration twist from body frame to inertial frame is given by
$${\rm{Ad}}_{\boldsymbol{g}}= \left(\begin{matrix}
	\boldsymbol{R}&\boldsymbol{0}_{3\times3}\\\tilde{\boldsymbol{u}}\boldsymbol{R}&\boldsymbol{R}
\end{matrix}\right)\in \mathbb{R}^{6\times6}.$$
\section*{Acknowledgment}

\nocite{1}  
\bibliographystyle{IEEEtran}
\bibliography{contact}

\end{document}